# How To Make the Gradients Small Stochastically: Even Faster Convex and Nonconvex SGD

(version 2)

Zeyuan Allen-Zhu
zeyuan@csail.mit.edu
Microsoft Research AI

January 6, 2018[*]

**Abstract**

Stochastic gradient descent (SGD) gives an optimal convergence rate when minimizing convex stochastic objectives $f(x)$. However, in terms of making the gradients small, the original SGD does not give an optimal rate, even when $f(x)$ is convex.

If $f(x)$ is convex, to find a point with gradient norm $\varepsilon$, we design an algorithm SGD3 with a near-optimal rate $\widetilde{O}(\varepsilon^{-2})$, improving the best known rate $O(\varepsilon^{-8/3})$ of [17].

If $f(x)$ is nonconvex, to find its $\varepsilon$-approximate local minimum, we design an algorithm SGD5 with rate $\widetilde{O}(\varepsilon^{-3.5})$, where previously SGD variants only achieve $\widetilde{O}(\varepsilon^{-4})$ [6, 15, 32]. This is no slower than the best known stochastic version of Newton's method in all parameter regimes [29].

---

# 1 Introduction

In convex optimization and machine learning, the classical goal is to design algorithms to decrease objective values, that is, to find points $x$ with $f(x) - f(x^*) \leq \varepsilon$. In contrast, the *rate of convergence for the gradients*, that is, the number of iterations $T$ needed to find a point $x$ with $\|\nabla f(x)\| \leq \varepsilon$, is a harder problem and sometimes needs new algorithmic ideas [26]. For instance, in the full-gradient setting, accelerated gradient descent alone is suboptimal for this new goal, and one needs additional tricks to get the fastest rate [26]. We review these tricks in Section 1.1.

In the convex (online) *stochastic* optimization, to the best of our knowledge, tight bounds are not yet known for finding points with small gradients. The best recorded rate was $T \propto \varepsilon^{-8/3}$ [17], and it was raised as an open question [1] regarding how to improve it.

In this paper, we design two new algorithms, SGD2 which gives rate $T \propto \varepsilon^{-5/2}$ using Nesterov's tricks, and SGD3 which gives an even better rate $T \propto \varepsilon^{-2} \log^3 \frac{1}{\varepsilon}$ which is optimal up to log factors.

We also apply our techniques to design SGD4 and SGD5 for non-convex optimization tasks.

**Motivation.** Studying the rate of convergence for the minimizing gradients can be important at least for the following two reasons.

- In many situations, points with small gradients fit better our final goals.

  Nesterov [26] considers the dual approach for solving constrained minimization problems. He argued that "the gradient value $\|\nabla f(x)\|$ serves as the measure of feasibility and optimality of the primal solution," and thus is the better goal for minimization purpose.[1]

  In matrix scaling [8, 11], given a non-negative matrix, one wants to re-scale its rows and columns to make it doubly stochastic. This problem has been applied in image reconstruction, operations research, decision and control, and other scientific disciplines (see survey [20]). The goal for matrix scaling is to find points with small gradients, but not small objectives.[2]

- Designing algorithms to find points with small gradients can help us understand non-convex optimization better and design faster non-convex machine learning algorithms.

  Without strong assumptions, non-convex optimization theory is always in terms of finding points with small gradients (i.e., approximate stationary points or local minima). Therefore, to understand non-convex stochastic optimization better, perhaps we should first figure out the best rate for *convex* stochastic optimization. In addition, if new algorithmic ideas are needed, can we also apply them to the non-convex world? We find positive answers to this question, and also obtain better rates for standard non-convex optimization tasks.

## 1.1 Review: Prior Work on Deterministic Convex Optimization

For convex optimization, Nesterov [26] discussed the difference between convergence for *objective values* vs. for *gradients*, and introduced two algorithms. We review his results as follows.

---

[1]Nesterov [26] studied $\min_{y \in Q}\{g(y) : Ay = b\}$ with convex $Q$ and strongly convex $g(y)$. The dual problem is $\min_x\{f(x)\}$ where $f(x) \stackrel{\text{def}}{=} \min_{y \in Q}\{g(y) + \langle x, b - Ay \rangle\}$. Let $y^*(x) \in Q$ be the (unique) minimizer of the internal problem, then $g(y^*(x)) - f(x) = \langle x, \nabla f(x) \rangle \leq \|x\| \cdot \|\nabla f(x)\|$.

[2]In matrix scaling, given a non-negative matrix $A \in \mathbb{R}^{n \times m}$, we want to find positive diagonal matrices $X \in \mathbb{R}^{n \times n}$, $Y \in \mathbb{R}^{m \times m}$ such that $XAY$ is close to being doubly-stochastic. There are several ways to define a convex objective $f(\cdot)$ for this problem. For instance, $f(x, y) = \sum_{i,j} A_{i,j} e^{x_i - y_j} - \mathbb{1}^\top x + \mathbb{1}^\top y$ in [11] and $f(y) = \sum_i \log(\sum_j A_{i,j} e^{y_j}) - \mathbb{1}^\top y$ in [8]. In these cases, "how close $XAY$ is to being doubly stochastic" is captured by $\|\nabla f(x)\| \leq \varepsilon$ as opposed to the objective value.

Suppose $f(x)$ is a Lipschitz smooth convex function with smoothness parameter $L$. Then, it is well-known that accelerated gradient descent (AGD) [24, 25] finds a point $x$ satisfying $f(x) - f(x^*) \leq \delta$ using $T = O(\frac{\sqrt{L}}{\sqrt{\delta}})$ gradient computations of $\nabla f(x)$. To turn this into a gradient guarantee, we can apply the smoothness property of $f(x)$ which gives $\|\nabla f(x)\|^2 \leq L(f(x) - f(x^*))$. This means

- to get a point $x$ with $\|\nabla f(x)\| \leq \varepsilon$, AGD converges in rate $T \propto \frac{L}{\varepsilon}$.

Nesterov [26] proposed two different tricks to improve upon such rate.

**Nesterov's First Trick: GD After AGD.** Recall that starting from a point $x_0$, if we perform $T$ steps of gradient descent (GD) $x_{t+1} = x_t - \frac{1}{L}\nabla f(x_t)$, then it satisfies $\sum_{t=0}^{T-1} \|\nabla f(x_t)\|^2 \leq L(f(x_0) - f(x^*))$ (see for instance [7, 22]). In addition, if this $x_0$ is already the output of AGD for another $T$ iterations, then it satisfies $f(x_0) - f(x^*) \leq O(\frac{L}{T^2})$. Putting the two inequalities together, we have $\min_{t=0}^{T-1} \{\|\nabla f(x_t)\|^2\} \leq O(\frac{L^2}{T^3})$. We call this method "GD after AGD," and it satisfies

- to get a point $x$ with $\|\nabla f(x)\| \leq \varepsilon$, "GD after AGD" converges in rate $T \propto \frac{L^{2/3}}{\varepsilon^{2/3}}$.

**Nesterov's Second Trick: AGD After Regularization.** Alternatively, we can also regularize $f(x)$ by defining $g(x) = f(x) + \frac{\sigma}{2}\|x - x_0\|^2$. This new function $g(x)$ is $\sigma$-strongly convex, so AGD converges *linearly*, meaning that using $T \propto \frac{\sqrt{L}}{\sqrt{\sigma}} \log \frac{L}{\varepsilon}$ gradients we can find a point $x$ satisfying $\|\nabla g(x)\|^2 \leq L(g(x) - g(x^*)) \leq \varepsilon^2$. If we choose $\sigma \propto \varepsilon$, then this implies $\|\nabla f(x)\| \leq \|\nabla g(x)\| + \varepsilon \leq 2\varepsilon$. We call this method "AGD after regularization," and it satisfies

- to get a point $x$ with $\|\nabla f(x)\| \leq \varepsilon$, "AGD after regularization" converges in rate $T \propto \frac{L^{1/2}}{\varepsilon^{1/2}} \log \frac{L}{\varepsilon}$.

**Nesterov's Lower Bound.** Recall that Nesterov constructed hard-instance functions $f(x)$ so that, when dimension is sufficiently high, first-order methods require at least $T = \Omega(\sqrt{L/\delta})$ computations of $\nabla f(x)$ to produce a point $x$ satisfying $f(x) - f(x^*) \leq \delta$ (see his textbook [24]). Since $f(x) - f(x^*) \leq \langle \nabla f(x), x - x^* \rangle \leq \|\nabla f(x)\| \cdot \|x - x^*\|$, this also implies a lower bound $T = \Omega(\sqrt{L/\varepsilon})$ to find a point $x$ with $\|\nabla f(x)\| \leq \varepsilon$. In other words,

- to get a point $x$ with $\|\nabla f(x)\| \leq \varepsilon$, "AGD after regularization" is optimal (up to a log factor).

## 1.2 Our Results: Stochastic Convex Optimization

Consider the stochastic setting where the convex objective $f(x) \stackrel{\text{def}}{=} \mathbb{E}_i[f_i(x)]$ and the algorithm can only compute stochastic gradients $\nabla f_i(x)$ at any point $x$ for a random $i$. Let $T$ be the number of stochastic gradient computations. It is well-known that stochastic gradient descent (SGD) finds a point $x$ with $f(x) - f(x^*) \leq \delta$ in (see for instance textbooks [9, 18, 28])

$$T = O\Big(\frac{\mathcal{V}}{\delta^2}\Big) \text{ iterations } \quad \text{or} \quad T = O\Big(\frac{\mathcal{V}}{\sigma\delta}\Big) \text{ if } f(x) \text{ is } \sigma\text{-strongly convex.}$$

Both rates are asymptotically optimal in terms of decreasing objective, and $\mathcal{V}$ is an absolute bound on the variance of the stochastic gradients. Using the same argument $\|\nabla f(x)\|^2 \leq L(f(x) - f(x^*))$ as before, SGD finds a point $x$ with $\|\nabla f(x)\| \leq \varepsilon$ in

$$T = O\Big(\frac{L^2 \mathcal{V}}{\varepsilon^4}\Big) \text{ iterations } \quad \text{or} \quad T = O\Big(\frac{L\mathcal{V}}{\sigma\varepsilon^2}\Big) \text{ if } f(x) \text{ is } \sigma\text{-strongly convex.} \quad \text{(SGD)}$$

These rates are not optimal. We investigate three approaches to improve such rates.

**New Approach 1: SGD after SGD.** Recall in Nesterov's first trick, he replaced the use of the inequality $\|\nabla f(x)\|^2 \leq L(f(x) - f(x^*))$ by $T$ steps of gradient descent. In the stochastic setting, can we replace this inequality with $T$ steps of SGD? We call this algorithm SGD1 and prove that

| | algorithm | | gradient complexity $T$ | | 2nd-order smooth |
|---|---|---|---|---|---|
| online convex | SGD | (naive) | $O(\varepsilon^{-4})$ | (folklore, see Theorem 4.2) | |
| | SGD1 | (SGD after SGD) | $O(\varepsilon^{-8/3})$ | (see [17] or Theorem 1) | |
| | SGD2 | (SGD after regularization) | $O(\varepsilon^{-5/2})$ | (see Theorem 2) | |
| | SGD3 | (SGD + recursive regularization) | $O(\varepsilon^{-2} \cdot \log^3 \frac{1}{\varepsilon})$ | (see Theorem 3) | |
| online strongly convex | SGD$^{\mathsf{sc}}$ | (naive) | $O(\varepsilon^{-2} \cdot \kappa)$ | (see Theorem 4.2) | no |
| | SGD1$^{\mathsf{sc}}$ | (SGD after SGD) | $O(\varepsilon^{-2} \cdot \kappa^{1/2})$ | (see Theorem 1) | |
| | SGD3$^{\mathsf{sc}}$ | (SGD + recursive regularization) | $O(\varepsilon^{-2} \cdot \log^3 \kappa)$ | (see Theorem 3) | |
| online nonconvex ($\sigma$-nonconvex) | SGD | (naive) | $O(\varepsilon^{-4})$ | (folklore, see e.g. [3, 17]) | |
| | SCSG | | $O(\varepsilon^{-10/3})$ | (see [21]) | |
| | SGD4 | | $O(\varepsilon^{-2} + \sigma\varepsilon^{-4})$ | (see Theorem 4) | |
| | Natasha1.5 | | $O(\varepsilon^{-3} + \sigma^{1/3}\varepsilon^{-10/3})$ | (see [3]) | |
| | SGD variants | | $\widetilde{O}(\varepsilon^{-4})$ | (see [6, 15, 32]) | |
| | SGD5 | | $\widetilde{O}(\varepsilon^{-3.5})$ | (see Theorem 5) | needed |
| | cubic Newton | | $\widetilde{O}(\varepsilon^{-3.5})$ | (see [29]) | |
| | Natasha2 | | $O(\varepsilon^{-3.25})$ | (see [3]) | |

Table 1: Comparison of first-order **online stochastic** methods for finding $\|\nabla f(x)\| \leq \varepsilon$. Following tradition, in these bounds, we hide variance and smoothness parameters in big-$O$ and only show the dependency on $\varepsilon$, the condition number $\kappa = \frac{L}{\sigma} \geq 1$ (if the objective is $\sigma$-strongly convex), or the nonconvexity parameter $\sigma$.

---

**Theorem 1** (informal). *For convex stochastic optimization, SGD1 finds $x$ with $\|\nabla f(x)\| \leq \varepsilon$ in*

$$T = O\Big(\frac{L^{2/3}\mathcal{V}}{\varepsilon^{8/3}}\Big) \text{ iterations} \quad or \quad T = O\Big(\frac{L^{1/2}\mathcal{V}}{\sigma^{1/2}\varepsilon^2}\Big) \text{ if } f(x) \text{ is } \sigma\text{-strongly convex.} \quad \text{(SGD1)}$$

We prove Theorem 1 in the general language of composite function minimization. This allows us to support an additional "proximal" term $\psi(x)$ and minimize $\psi(x) + f(x)$. For instance, if $\psi(x) = 0$ if $x \in Q$ and $\psi(x) = +\infty$ if $x \notin Q$ for some convex $Q$, then Theorem 1 is to minimize $f(x)$ over $Q$.

The rate $T \propto \varepsilon^{-8/3}$, in the special case of $\psi(x) \equiv 0$, was first recorded by Ghadimi and Lan [17]. Their algorithm is more involved because they also attempted to tighten the lower order terms using acceleration. To the best of our knowledge, our rate $T \propto \frac{1}{\sigma^{1/2}\varepsilon^2}$ in Theorem 1 is new.

**New Approach 2: SGD after regularization.** Recall that in Nesterov's second trick, he defined $g(x) = f(x) + \frac{\sigma}{2}\|x - x_0\|^2$ as a regularized version of $f(x)$, and applied the strongly-convex version of AGD to minimize $g(x)$. Can we apply this trick to the stochastic setting?

Note the parameter $\sigma$ has to be on the magnitude of $\varepsilon$ because $\nabla g(x) = \nabla f(x) + \sigma(x - x_0)$ and we wish to make sure $\|\nabla f(x)\| = \|\nabla g(x)\| \pm \varepsilon$. Therefore, if we apply SGD1 to minimize $g(x)$ to find a point $\|\nabla g(x)\| \leq \varepsilon$, the convergence rate is $T \propto \frac{1}{\sigma^{1/2}\varepsilon^2} = \frac{1}{\varepsilon^{2.5}}$. We call this algorithm SGD2.

**Theorem 2** (informal). *For convex stochastic optimization, SGD2 finds $x$ with $\|\nabla f(x)\| \leq \varepsilon$ in*

$$T = O\Big(\frac{L^{1/2}\mathcal{V}}{\varepsilon^{5/2}}\Big) \text{ iterations.} \quad \text{(SGD2)}$$

We prove Theorem 2 also in the general proximal language. This $T \propto \varepsilon^{-5/2}$ rate improves the best known result of $T \propto \varepsilon^{-8/3}$, but is still far from the lower bound $\Omega(\mathcal{V}/\varepsilon^2)$.

**New Approach 3: SGD and recursive regularization.** In the second approach above, the $\varepsilon^{0.5}$ sub-optimality gap is due to the choice of $\sigma \propto \varepsilon$ which ensures $\|\sigma(x - x_0)\| \leq \varepsilon$.

Intuitively, if $x_0$ were sufficiently close to $x^*$ (and thus were also close to the approximate minimizer $x$), then we could choose $\sigma \gg \varepsilon$ so that $\|\sigma(x - x_0)\| \leq \varepsilon$ still holds. In other words, an appropriate *warm start* $x_0$ could help us break the $\varepsilon^{-2.5}$ barrier and get a better convergence rate. However, how to find such $x_0$? We find it by constructing a "less warm" starting point and so on. This process is summarized by the following algorithm which recursively finds the warm starts.

Starting from $f^{(0)}(x) \stackrel{\text{def}}{=} f(x)$, we define $f^{(s)}(x) \stackrel{\text{def}}{=} f^{(s-1)}(x) + \frac{\sigma_s}{2}\|x - \widehat{x}_s\|^2$ where $\sigma_s = 2\sigma_{s-1}$ and $\widehat{x}_s$ is an approximate minimizer of $f^{(s-1)}(x)$ that is simply calculated from the naive SGD. We call this method SGD3, and prove that

**Theorem 3** (informal). *For convex stochastic optimization, SGD3 finds $x$ with $\|\nabla f(x)\| \leq \varepsilon$ in*

$$T = O\Big(\frac{\log^3(L/\varepsilon) \cdot \mathcal{V}}{\varepsilon^2}\Big) \text{ iterations} \quad \text{or} \quad T = O\Big(\frac{\log^3(L/\sigma) \cdot \mathcal{V}}{\varepsilon^2}\Big) \text{ if } f(x) \text{ is } \sigma\text{-strongly convex.} \tag{SGD3}$$

Our new rates in Theorem 3 not only improve the best known result of $T \propto \varepsilon^{-8/3}$, but also are near optimal because $\Omega(\mathcal{V}/\varepsilon^2)$ is clearly a lower bound: even to decide whether a point $x$ has $\|\nabla f(x)\| \leq \varepsilon$ or $\|\nabla f(x)\| > 2\varepsilon$ requires $\Omega(\mathcal{V}/\varepsilon^2)$ samples of the stochastic gradient.[3]

Perhaps interestingly, our dependence on the smoothness parameter $L$ (or the condition number $\kappa \stackrel{\text{def}}{=} L/\sigma$ if strongly convex) is only polylogarithmic, as opposed to polynomial in all previous results.

## 1.3 Our Applications: Stochastic Non-Convex Optimization

One natural question to ask is whether our techniques for convex stochastic optimization translate to non-convex performance guarantees? We design two SGD variants to tackle this question.

**New Approach 4: SGD for stationary points.** In the first application, we minimize a nonconvex stochastic function $f(x) = \mathbb{E}_i[f_i(x)]$ that is $L$-smooth and of $\sigma$-bounded nonconvexity (or $\sigma$-nonconvex for short), meaning that all eigenvalues of $\nabla^2 f(x)$ are above $-\sigma$ for some parameter $\sigma \in [0, L]$. Our goal is to again to find $x$ with $\|\nabla f(x)\| \leq \varepsilon$. To solve this task, we recursively minimize $g(x) = f(x) + \sigma\|x - \widehat{x}_s\|$ which is a $\sigma$-strongly convex, and let the resulting point be $\widehat{x}_{s+1}$.

Such recursive regularization techniques for non-convex optimization have appeared in prior works [3, 10]. However, different from them, we only use simple SGD variants to minimize each $g(x)$ and then use SGD3 to get small gradient. We call this algorithm SGD4 and prove that

**Theorem 4** (informal). *For non-convex stochastic optimization with $\sigma$-bounded nonconvexity, SGD4 finds $x$ with $\|\nabla f(x)\| \leq \varepsilon$ in*

$$T = O\Big(\frac{L(f(x_0) - f(x^*)) + \mathcal{V}\log^3(1/\varepsilon)}{\varepsilon^2} + \frac{\sigma\mathcal{V}(f(x_0) - f(x^*))}{\varepsilon^4}\Big) \text{ iterations} \tag{SGD4}$$

Perhaps surprisingly, this simple SGD variant already outperforms previous results in the regime of $\sigma \leq \varepsilon L$. We closely compare SGD4 to them in Figure 1(a) and Table 1.

**New Approach 5: SGD for local minima.** In the second application, we tackle the more ambitious goal of finding a point $x$ with both $\|\nabla f(x)\| \leq \varepsilon$ and $\nabla^2 f(x) \succeq -\delta \mathbf{I}$, known as an $(\varepsilon, \delta)$-*approximate local minimum*. For this harder task, one needs the following two standard assumptions: each $f_i(x)$ is $L$-smooth and $f(x)$ is $L_2$-second-order smooth. (The later means $\|\nabla^2 f(x) - \nabla^2 f(y)\|_2 \leq L_2\|x - y\|$ for every $x, y$.)

---
[3]Indeed, $f(x) = \mathbb{E}_i[f_i(x)]$ and it satisfies $\mathbb{E}_i\big[\|\nabla f(x) - \nabla f_i(x)\|^2\big] \leq \mathcal{V}$. If we sample $N$ i.i.d. indices $i_1, \ldots, i_N$, then $\mathbb{E}_{i_1,\ldots,i_N}\big[\|\nabla f(x) - \frac{1}{N}\big(\nabla f_{i_1}(x) + \cdots + \nabla f_{i_N}(x)\big)\|^2\big] \leq \frac{\mathcal{V}}{N}$.

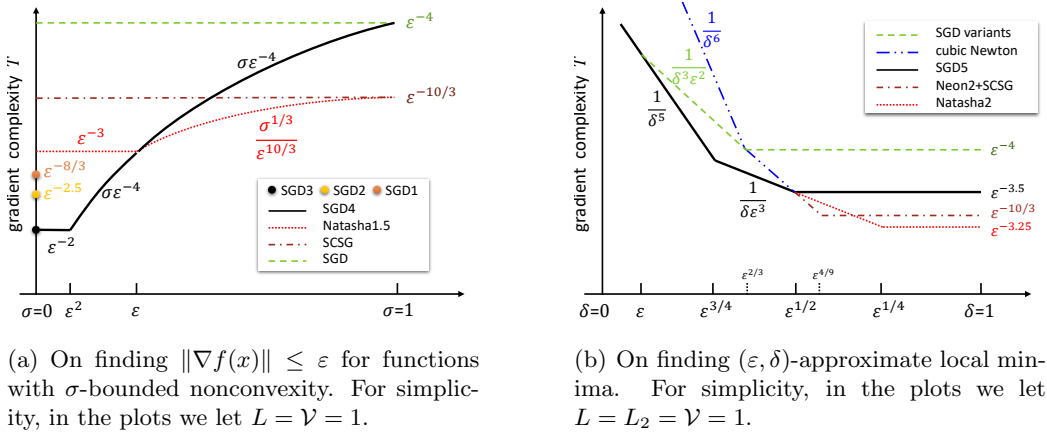

(a) On finding $\|\nabla f(x)\| \leq \varepsilon$ for functions with $\sigma$-bounded nonconvexity. For simplicity, in the plots we let $L = \mathcal{V} = 1$.

(b) On finding $(\varepsilon, \delta)$-approximate local minima. For simplicity, in the plots we let $L = L_2 = \mathcal{V} = 1$.

Figure 1: Comparison on convergence rates for online stochastic nonconvex optimization

Motivated by the "swing by saddle point" framework of [3], we combine SGD variants with Oja's algorithm of [5] to design a new algorithm SGD5.[4] We prove that

**Theorem 5** (informal). *For non-convex stochastic optimization, SGD5 finds $x$ with $\|\nabla f(x)\| \leq \varepsilon$ and $\nabla^2 f(x) \succeq -\delta \mathbf{I}$ in (ignoring the dependency on $L, L_2, \mathcal{V}, f(x_0) - f(x^*)$ for simplicity)*

$$T = \widetilde{O}\Big(\frac{1}{\delta^5} + \frac{1}{\varepsilon^{3.5}} + \frac{1}{\delta\varepsilon^3}\Big) \text{ iterations} \tag{SGD5}$$

We compare SGD5 to known results in Figure 1(b). Perhaps surprisingly, our SGD5, being a simple SGD variant, performs no worse than cubic regularized Newton's method with $T = \widetilde{O}\big(\frac{1}{\varepsilon^{3.5}} + \frac{1}{\delta^6} + \frac{1}{\varepsilon^2 \delta^3}\big)$ [29] or the best known SGD variant with $T = \widetilde{O}\big(\frac{1}{\varepsilon^4} + \frac{1}{\delta^5} + \frac{1}{\varepsilon^2 \delta^3}\big)$ [6]. Only when $\sigma > \sqrt{\varepsilon}$, SGD5 is outperformed by variance-reduction based methods Neon2+SCSG [6] and Natasha2 [3].

*Remark* 1.1. Existing SGD variants to find approximate local minima are all based on the "escape saddle points" approach. In contrast, SGD5 is based on the alternative "swing by saddle point" approach. For the difference between the two, we refer interested readers to [3, 6].

### 1.4 Roadmap

We introduce notions in Section 2 and formalize the convex problem in Section 3. We review classical (convex) SGD theorems with objective decrease in Section 4. We give an auxiliary lemma in Section 5 show our SGD3 results in Section 6. We apply our techniques to non-convex optimization and give algorithms SGD4 and SGD5 in Section 7. We discuss more related work in Appendix A, and show our results on SGD1 and SGD2 respectively in Appendix B and Appendix C.

## 2 Preliminaries

Throughout this paper, we denote by $\|\cdot\|$ the Euclidean norm. We use $i \in_R [n]$ to denote that $i$ is generated from $[n] = \{1, 2, \ldots, n\}$ uniformly at random. We denote by $\nabla f(x)$ the gradient of function $f$ if it is differentiable, and $\partial f(x)$ any subgradient if $f$ is only Lipschitz continuous. We denote by $\mathbb{I}[event]$ the indicator function of probabilistic events.

---

[4]Oja's algorithm [27] is itself an SGD variant of power method to find approximate eigenvectors. We rely on the recent work [5] which gives the optimal rate for Oja's algorithm.

We denote by $\|\mathbf{A}\|_2$ the spectral norm of matrix $\mathbf{A}$. For symmetric matrices $\mathbf{A}$ and $\mathbf{B}$, we write $\mathbf{A} \succeq \mathbf{B}$ to indicate that $\mathbf{A} - \mathbf{B}$ is positive semidefinite (PSD). Therefore, $\mathbf{A} \succeq -\sigma \mathbf{I}$ if and only if all eigenvalues of $\mathbf{A}$ are no less than $-\sigma$. We denote by $\lambda_{\min}(\mathbf{A})$ and $\lambda_{\max}(\mathbf{A})$ the minimum and maximum eigenvalue of a symmetric matrix $\mathbf{A}$.

Recall some definitions on strong convexity and smoothness (and they have other equivalent definitions, see textbook [24]).

**Definition 2.1.** *For a function $f \colon \mathbb{R}^d \to \mathbb{R}$,*
- *$f$ is $\sigma$-**strongly convex** if $\forall x, y \in \mathbb{R}^d$, it satisfies $f(y) \geq f(x) + \langle \partial f(x), y - x \rangle + \frac{\sigma}{2}\|x - y\|^2$.*
- *$f$ is of $\sigma$-bounded nonconvexity (or $\sigma$-**nonconvex** for short) if $\forall x, y \in \mathbb{R}^d$, it satisfies $f(y) \geq f(x) + \langle \partial f(x), y - x \rangle - \frac{\sigma}{2}\|x - y\|^2$.* [5]
- *$f$ is $L$-Lipschitz smooth (or $L$-**smooth** for short) if $\forall x, y \in \mathbb{R}^d$, $\|\nabla f(x) - \nabla f(y)\| \leq L\|x - y\|$.*
- *$f$ is $L_2$-**second-order smooth** if $\forall x, y \in \mathbb{R}^d$, it satisfies $\|\nabla^2 f(x) - \nabla^2 f(y)\|_2 \leq L_2 \|x - y\|$.*

**Definition 2.2.** *For composite function $F(x) = \psi(x) + f(x)$ where $\psi(x)$ is proper convex, given a parameter $\eta > 0$, the **gradient mapping** of $F(\cdot)$ at point $x$ is*

$$\mathcal{G}_{F,\eta}(x) \stackrel{\text{def}}{=} \frac{1}{\eta}(x - x^+) \qquad \text{where} \qquad x^+ = \arg\min_y \left\{ \psi(y) + \langle \nabla f(x), y \rangle + \frac{1}{2\eta}\|y - x\|^2 \right\}$$

*In particular, if $\psi(\cdot) \equiv 0$, then $\mathcal{G}_{F,\eta}(x) \equiv \nabla f(x)$.*

Recall the following property about gradient mapping —see for instance [31, Lemma 3.7])

**Lemma 2.3.** *Let $F(x) = \psi(x) + f(x)$ where $\psi(x)$ is proper convex and $f(x)$ is $\sigma$-strongly convex and $L$-smooth. For every $x, y \in \{x \in \mathbb{R}^d \colon \psi(x) < +\infty\}$, letting $x^+ = x - \eta \cdot \mathcal{G}_{F,\eta}(x)$, we have*

$$\forall \eta \in \left(0, \frac{1}{L}\right] \colon \quad F(y) \geq F(x^+) + \langle \mathcal{G}_{F,\eta}(x), y - x \rangle + \frac{\eta}{2}\|\mathcal{G}_{F,\eta}(x)\|^2 + \frac{\sigma}{2}\|y - x\|^2 .$$

The following definition and properties of Fenchel dual for convex functions is classical, and can be found for instance in the textbook [28].

**Definition 2.4.** *Given proper convex function $h(y)$, its Fenchel dual $h^*(\beta) \stackrel{\text{def}}{=} \max_y \{y^\top \beta - h(y)\}$.*

**Proposition 2.5.** $\nabla h^*(\beta) = \arg\max_y \{y^\top \beta - h(y)\}$.

**Proposition 2.6.** *If $h(\cdot)$ is $\sigma$-strongly convex, then $h^*(\cdot)$ is $\frac{1}{\sigma}$-smooth.*

## 3 Problem Formalization

Throughout this paper (except our nonconvex application Section 7), we minimize the following *convex* stochastic composite objective:

$$\min_{x \in \mathbb{R}^d} \left\{ F(x) = \psi(x) + f(x) \stackrel{\text{def}}{=} \psi(x) + \frac{1}{n}\sum_{i \in [n]} f_i(x) \right\} , \tag{3.1}$$

where

1. $\psi(x)$ is proper convex (a.k.a. the proximal term),
2. $f_i(x)$ is differentiable for every $i \in [n]$,
3. $f(x)$ is $L$-smooth and $\sigma$-strongly convex for some $\sigma \in [0, L]$ that could be zero,
4. $n$ can be very large of even infinite (so $f(x) = \mathbb{E}_i[f_i(x)]$),[6] and

---
[5]Previous authors also refer to this notion as "approximate convex", "almost convex", "hypo-convex", "semi-convex", or "weakly-convex." We call it $\sigma$-nonconvex to stress the point that $\sigma$ can be as large as $L$ (any $L$-smooth function is automatically $L$-nonconvex).

[6]All of the results in this paper apply to the case when $n$ is infinite, because we focus on online methods. However, we still introduce $n$ to simplify notations.

**Algorithm 1** $\texttt{SGD}(F, x_0, \alpha, T)$

**Input:** function $F(x) = \psi(x) + \frac{1}{n}\sum_{i=1}^{n} f_i(x)$; initial vector $x_0$; learning rate $\alpha > 0$; $T \geq 1$.
  ⋄ *if $f(x) = \frac{1}{n}\sum_{i=1}^{n} f_i(x)$ is L-smooth, optimal choice $\alpha = \Theta\big(\min\big\{\frac{\|x_0 - x^*\|}{\sqrt{\mathcal{V}T}}, \frac{1}{L}\big\}\big)$*

1: **for** $t = 0$ to $T - 1$ **do**
2:     $i \leftarrow$ a random index in $[n]$;
3:     $x_{t+1} \leftarrow \arg\min_{y \in \mathbb{R}^d}\{\psi(y) + \frac{1}{2\alpha}\|y - x_t\|^2 + \langle \nabla f_i(x_t), y \rangle\}$;
4: **return** $\overline{x} = \frac{x_1 + \cdots + x_T}{T}$.

---

**Algorithm 2** $\texttt{SGD}^{\textsf{sc}}(F, x_0, \sigma, L, T)$

**Input:** function $F(x) = \psi(x) + \frac{1}{n}\sum_{i=1}^{n} f_i(x)$; initial vector $x_0$; parameters $0 < \sigma \leq L$; $T \geq \frac{L}{\sigma}$.
  ⋄ *$f(x)$ is $\sigma$-strongly convex and $f(x) = \frac{1}{n}\sum_{i=1}^{n} f_i(x)$ is L-smooth*

1: **for** $t = 1$ to $N = \lfloor \frac{T}{8L/\sigma} \rfloor$ **do**    $x_t \leftarrow \texttt{SGD}\big(F, x_{t-1}, \frac{1}{2L}, \frac{4L}{\sigma}\big)$;
2: **for** $k = 1$ to $K = \lfloor \log_2(\sigma T / 16L) \rfloor$ **do**    $x_{N+k} \leftarrow \texttt{SGD}\big(F, x_{N+k-1}, \frac{1}{2^k L}, \frac{2^{k+2}L}{\sigma}\big)$;
3: **return** $\overline{x} = x_{N+K}$.

---

5. the stochastic gradients $\nabla f_i(x)$ have a bounded variance (over the domain of $\psi(\cdot)$), that is

$$\forall x \in \{y \in \mathbb{R}^d \,|\, \psi(y) < +\infty\}: \quad \mathbb{E}_{i \in_R [n]} \|\nabla f(x) - \nabla f_i(x)\|^2 \leq \mathcal{V} \ .$$

We emphasize that the above assumptions are all classical.

In the rest of the paper, we define $T$, the gradient complexity, as the number of computations of $\nabla f_i(x)$. We search for points $x$ so that the gradient mapping $\|\mathcal{G}_{F,\eta}(x)\| \leq \varepsilon$ for any $\eta \approx \frac{1}{L}$. Recall from Definition 2.2 that if there is no proximal term (i.e., $\psi(x) \equiv 0$), then $\mathcal{G}_{F,\eta}(x) = \nabla f(x)$ for any $\eta > 0$. We want to study the best tradeoff between the gradient complexity $T$ and the error $\varepsilon$.

We say an algorithm is *online* if its gradient complexity $T$ is independent of $n$. This tackles the big-data scenario when $n$ is extremely large or even infinite (i.e., $f(x) = \mathbb{E}_i[f_i(x)]$ for some random variable $i$). The stochastic gradient descent (SGD) method and all of its variants studied in this paper are online. In contrast, GD, AGD [24, 25], and Katyusha [2] are offline methods because their gradient complexity depends on $n$ (see Table 2 in appendix).

## 4 Review: SGD with Objective Value Convergence

Recall that stochastic gradient descent (SGD) repeatedly performs *proximal updates* of the form

$$x_{t+1} = \arg\min_{y \in \mathbb{R}^d}\{\psi(y) + \frac{1}{2\alpha}\|y - x_t\|^2 + \langle \nabla f_i(x_t), y \rangle\} \ ,$$

where $\alpha > 0$ is some learning rate, and $i$ is chosen in $1, 2, \ldots, n$ uniformly at random per iteration. Note that if $\psi(y) \equiv 0$ then $x_{t+1} = x_t - \alpha \nabla f_i(x_t)$. For completeness' sake, we summarize it in Algorithm 1. If $f(x)$ is also known to be strongly convex, to get the tightest convergence rate, one can repeatedly apply SGD with decreasing learning rate $\alpha$ [19]. We summarize this algorithm as $\texttt{SGD}^{\textsf{sc}}$ in Algorithm 2.

The following theorem describes the rates of convergence in objective values for SGD and $\texttt{SGD}^{\textsf{sc}}$ respectively. Their proofs are classical (and included in Appendix D); however, for our exact statements, we cannot find them recorded anywhere.[7]

**Theorem 4.1.** *Let $x^* \in \arg\min_x\{F(x)\}$. To solve Problem (3.1) given a starting vector $x_0 \in \mathbb{R}^d$,*

(a) *$\mathtt{SGD}(F, x_0, \alpha, T)$ outputs $\overline{x}$ satisfying $\mathbb{E}[F(\overline{x})] - F(x^*) \leq \frac{\alpha \mathcal{V}}{2(1-\alpha L)} + \frac{\|x_0 - x^*\|^2}{2\alpha T}$ as long as $\alpha < 1/L$. In particular, if $\alpha$ is tuned optimally, it satisfies*
$$\mathbb{E}[F(\overline{x})] - F(x^*) \leq O\big(\tfrac{L\|x_0-x^*\|^2}{T} + \tfrac{\sqrt{\mathcal{V}}\|x_0-x^*\|}{\sqrt{T}}\big) \ .$$

(b) *If $f(x)$ is $\sigma$-strongly convex and $T \geq \frac{L}{\sigma}$, then $\mathtt{SGD}^{\mathsf{sc}}(F, x_0, \sigma, L, T)$ outputs $\overline{x}$ satisfying*
$$\mathbb{E}[F(\overline{x})] - F(x^*) \leq O\big(\tfrac{\mathcal{V}}{\sigma T}\big) + \big(1 - \tfrac{\sigma}{L}\big)^{\Omega(T)} \sigma \|x_0 - x^*\|^2 \ .$$

As a sanity check, if $\mathcal{V} = 0$, the convergence rate of SGD matches that of GD. (However, if $\mathcal{V} = 0$, one can apply accelerated gradient descent of Nesterov [23, 24] instead for a faster rate.)

To turn Theorem 4.1 into a rate of convergence for the gradients, we can simply apply Lemma 2.3 which implies
$$\forall \eta \in \big(0, \tfrac{1}{L}\big]: \quad \tfrac{\eta}{2}\|\mathcal{G}_{F,\eta}(\overline{x})\|^2 \leq F(\overline{x}) - F(\overline{x}^+) \leq F(\overline{x}) - F(x^*) \ . \tag{4.1}$$

**Theorem 4.2.** *Let $x^* \in \arg\min_x\{F(x)\}$. To solve Problem (3.1) given a starting vector $x_0 \in \mathbb{R}^d$ and any $\eta = \frac{C}{L}$ where $C \in (0, 1]$ is some absolute constant,*

(a) *$\mathtt{SGD}$ outputs $\overline{x}$ satisfying $\mathbb{E}[\|\mathcal{G}_{F,\eta}(\overline{x})\|^2] \leq O\big(\tfrac{L^2\|x_0-x^*\|^2}{T} + \tfrac{L\sqrt{\mathcal{V}}\|x_0-x^*\|}{\sqrt{T}}\big).$*

(b) *if $T \geq \frac{L}{\sigma}$, then $\mathtt{SGD}^{\mathsf{sc}}$ outputs $\overline{x}$ satisfying $\mathbb{E}[\|\mathcal{G}_{F,\eta}(\overline{x})\|^2] \leq O\big(\tfrac{L\mathcal{V}}{\sigma T}\big) + \big(1 - \tfrac{\sigma}{L}\big)^{\Omega(T)} \sigma L \|x_0-x^*\|^2.$*

**Corollary 4.3.** *Hiding $\mathcal{V}, L, \|x_0 - x^*\|$ in the big-O notation, classical SGD finds $x$ with*

$$F(x) - F(x^*) \leq O(T^{-1/2}) \quad \|\mathcal{G}_{F,\eta}(x)\| \leq O(T^{-1/4}) \quad \textit{for Problem (3.1), or}$$
$$F(x) - F(x^*) \leq O((\sigma T)^{-1}) \quad \|\mathcal{G}_{F,\eta}(x)\| \leq O((\sigma T)^{-1/2}) \quad \textit{if $f(\cdot)$ is $\sigma$-strongly convex for $\sigma > 0$.}$$

## 5 An Auxiliary Lemma on Regularization

Consider a regularized objective
$$G(x) \stackrel{\text{def}}{=} \psi(x) + g(x) \stackrel{\text{def}}{=} \psi(x) + \Big(f(x) + \sum_{s=1}^{S} \frac{\sigma_s}{2} \|x - \widehat{\mathsf{x}}_s\|^2\Big) \ , \tag{5.1}$$

where $\widehat{\mathsf{x}}_1, \ldots, \widehat{\mathsf{x}}_S$ are fixed vectors in $\mathbb{R}^d$. The following lemma says that, if we find an approximate stationary point $x$ of $G(x)$, then it is also an approximate stationary point of $F(x)$ up to some additive error.

**Lemma 5.1.** *Suppose $\psi(x)$ is proper convex and $f(x)$ is convex and $L$-smooth. By definition, $g(x)$ is $\widetilde{\sigma}$-strongly convex with $\widetilde{\sigma} \stackrel{\text{def}}{=} \sum_{s=1}^{S} \sigma_s$. Let $x^*$ be the unique minimizer of $G(y)$ in (5.1), and $x$ be an arbitrary vector in the domain of $\{x \in \mathbb{R}^d : \psi(x) < +\infty\}$. Then, for every $\eta \in \big(0, \frac{1}{L+\widetilde{\sigma}}\big]$, we*

---

[7]In the special case $\psi(x) \equiv 0$, Theorem 4.1(a) and 4.1(b) are folklore (see for instance [28]). If $\psi(x) \not\equiv 0$, Theorem 4.1(a) is recorded when $\psi(x)$ is Lipschitz or smooth [14], but we would not like to impose such assumptions. A variant of Theorem 4.1(b) is recorded for the accelerated version of SGD [16], but with a slightly worse rate $T = O\big(\frac{\mathcal{V}}{\sigma T} + \frac{L\|x_0-x^*\|^2}{T^2}\big)$. If the readers find either statement explicitly stated somewhere, please let us know and we would love to include appropriate citations.

have
$$\|\mathcal{G}_{F,\eta}(x)\| \leq \sum_{s=1}^{S} \sigma_s \|x^* - \widehat{\mathsf{x}}_s\| + 3\|\mathcal{G}_{G,\eta}(x)\| \ .$$

*Remark* 5.2. Lemma 5.1 should be easy to prove in the special case of $\psi(x) \equiv 0$. Indeed,

$$\|\nabla f(x)\| = \|\nabla g(x) + \sum_s \sigma_s(x - \widehat{\mathsf{x}}_s)\| \overset{\text{①}}{\leq} \|\nabla g(x)\| + \sum_s \sigma_s \|x - \widehat{\mathsf{x}}_s\|$$

$$\overset{\text{②}}{\leq} \|\nabla g(x)\| + \sum_s \sigma_s \|x^* - \widehat{\mathsf{x}}_s\| + \widetilde{\sigma}\|x^* - x\| \overset{\text{③}}{\leq} 2\|\nabla g(x)\| + \sum_s \sigma_s \|x^* - \widehat{\mathsf{x}}_s\| \ .$$

Above, inequalities ① and ② both use the triangle inequality; and inequality ③ is due to the $\widetilde{\sigma}$-strong convexity of $g(x)$ (see for instance [24, Sec. 2.1.3]).

*Proof of Lemma 5.1.* Define

$$z = \arg\min_y \left\{\psi(y) + \langle \nabla f(x), y\rangle + \frac{1}{2\eta}\|y - x\|^2\right\}$$

$$\overline{z} = \arg\min_y \left\{\psi(y) + \langle \nabla f(x) + \sum_{s=1}^{S} \sigma_s(x - \widehat{\mathsf{x}}_s), y\rangle + \frac{1}{2\eta}\|y - x\|^2\right\}$$

We have by definition $\mathcal{G}_{F,\eta}(x) = \frac{x-z}{\eta}$ and $\mathcal{G}_{G,\eta}(x) = \frac{x-\overline{z}}{\eta}$. Therefore, by triangle inequality,

$$\|\mathcal{G}_{F,\eta}(x)\| \leq \|\mathcal{G}_{G,\eta}(x)\| + \frac{1}{\eta}\|z - \overline{z}\| \ . \tag{5.2}$$

On the other hand, let us denote by $h(y) \overset{\text{def}}{=} \psi(y) + \frac{1}{2\eta}\|y\|^2$ and recall the definition of Fenchel dual $h^*(\beta) = \max_y\{y^\top \beta - h(y)\}$. Proposition 2.5 says $\nabla h^*(\beta) = \max_y\{y^\top \beta - h(y)\}$. This implies

$$z = \nabla h^*(\tfrac{x}{\eta} - \nabla f(x)) \quad \text{and} \quad \overline{z} = \nabla h^*(\tfrac{x}{\eta} - \nabla f(x) - \textstyle\sum_{s=1}^{S} \sigma(x - \widehat{\mathsf{x}}_s)) \ .$$

Using the property that $h^*(\cdot)$ is $\eta$-smooth (because $h(y)$ is $1/\eta$-strongly convex, see Proposition 2.6), we have

$$\tfrac{1}{\eta}\|z - \overline{z}\| \leq \|\textstyle\sum_{s=1}^{S} \sigma_s(x - \widehat{\mathsf{x}}_s)\| \leq \textstyle\sum_{s=1}^{S} \sigma_s\|x^* - \widehat{\mathsf{x}}_s\| + \widetilde{\sigma}\|x - x^*\| \ . \tag{5.3}$$

Next, recall the following property about gradient mapping (see Lemma 2.3 with $y = x^*$):[8]

$$\forall \eta \leq \frac{1}{L + \widetilde{\sigma}}: \quad G(x^*) \geq G(\overline{z}) + \langle \mathcal{G}_{G,\eta}(x), x^* - x\rangle + \frac{\eta}{2}\|\mathcal{G}_{G,\eta}(x)\|^2 + \frac{\widetilde{\sigma}}{2}\|x^* - x\|^2 \ .$$

Using $G(x^*) \leq G(\overline{z})$, the non-negativity of $\|\mathcal{G}_{G,\eta}(x)\|^2$, and Young's inequality $|\langle \mathcal{G}_{G,\eta}(x), x^* - x\rangle| \leq \frac{1}{\widetilde{\sigma}}\|\mathcal{G}_{G,\eta}(x)\|^2 + \frac{\widetilde{\sigma}}{4}\|x - x^*\|^2$, we have

$$\frac{\widetilde{\sigma}^2}{4}\|x - x^*\|^2 \leq \|\mathcal{G}_{G,\eta}(x)\|^2 \ . \tag{5.4}$$

Finally, combining (5.2), (5.3), and (5.4), we have the desired result. □

---
[8]To apply Lemma 2.3, we observe that $g(x) = f(x) + \sum_{s=1}^{S} \frac{\sigma_s}{2}\|x - \widehat{\mathsf{x}}_s\|^2$ is $\widetilde{\sigma}$-strongly convex and $(L + \widetilde{\sigma})$-smooth.

# 6  Approach 3: SGD and Recursive Regularization

In this section, add a logarithmic number of regularizers to the objective, each centered at a different but carefully chosen point. Specifically, given parameters $\sigma_1, \ldots, \sigma_S > 0$, we define functions

$$F^{(0)}(x) \stackrel{\text{def}}{=} F(x) \quad \text{and} \quad F^{(s)}(x) \stackrel{\text{def}}{=} F^{(s-1)}(x) + \frac{\sigma_s}{2}\|x - \widehat{x}_s\|^2 \quad \text{for } s = 1, 2, \ldots, S$$

where each $\widehat{x}_s$ (for $s \geq 1$) is an approximate minimizer of $F^{(s-1)}(x)$.

If $f(x)$ is $\sigma$-strongly convex, then we choose $S \approx \log_2 \frac{L}{\sigma}$ and let $\sigma_0 = \sigma$ and $\sigma_s = 2\sigma_{s-1}$. To calculate each $\widehat{x}_s$, we apply $\mathsf{SGD}^{\mathsf{sc}}$ for $\frac{T}{S}$ iterations. This totals to a gradient complexity of $T$. We summarize this method as $\mathsf{SGD3}^{\mathsf{sc}}$ in Algorithm 3.

If $f(x)$ is not strongly convex, then we regularize it by $G(x) = F(x) + \frac{\sigma}{2}\|x - x_0\|^2$ for some small parameter $\sigma > 0$, and then apply $\mathsf{SGD3}^{\mathsf{sc}}$. We summarize this final method as $\mathsf{SGD3}$ in Algorithm 4.

We prove the following main theorem:

---

**Theorem 3** (SGD3). *Let $x^* \in \arg\min_x\{F(x)\}$. To solve Problem (3.1) given a starting vector $x_0 \in \mathbb{R}^d$ and any $\eta = \frac{C}{L}$ for some absolute constant $C \in (0, 1]$.*

(a) *If $f(x)$ is $\sigma$-strongly convex for $\sigma \in (0, L]$ and $T \geq \frac{L}{\sigma} \log \frac{L}{\sigma}$, then $\mathsf{SGD3}^{\mathsf{sc}}(F, x_0, \sigma, L, T)$ outputs $\overline{x}$ satisfying*

$$\mathbb{E}[\|\mathcal{G}_{F,\eta}(\overline{x})\|] \leq O\Big(\frac{\sqrt{\mathcal{V}} \cdot \log^{3/2} \frac{L}{\sigma}}{\sqrt{T}}\Big) + \Big(1 - \frac{\sigma}{L}\Big)^{\Omega(T/\log(L/\sigma))} \sigma\|x_0 - x^*\| \ .$$

(b) *If $\sigma \in (0, L]$ and $T \geq \frac{L}{\sigma} \log \frac{L}{\sigma}$, then $\mathsf{SGD3}(F, x_0, \sigma, L, T)$ outputs $\overline{x}$ satisfying*

$$\mathbb{E}[\|\mathcal{G}_{F,\eta}(\overline{x})\|] \leq O\Big(\sigma\|x_0 - x^*\| + \frac{\sqrt{\mathcal{V}} \cdot \log^{3/2} \frac{L}{\sigma}}{\sqrt{T}}\Big) + \Big(1 - \frac{\sigma}{L}\Big)^{\Omega(T/\log(L/\sigma))} \sigma\|x_0 - x^*\| \ .$$

*If $\sigma$ is appropriately chosen, then we find $\overline{x}$ with $\mathbb{E}[\|\mathcal{G}_{F,\eta}(\overline{x})\|] \leq \varepsilon$ in gradient complexity*

$$T \leq O\Big(\frac{\mathcal{V} \cdot \log^3 \frac{L\|x_0 - x^*\|}{\varepsilon}}{\varepsilon^2} + \frac{L\|x_0 - x^*\|}{\varepsilon} \log \frac{L\|x_0 - x^*\|}{\varepsilon}\Big) \ .$$

---

*Remark* 6.1. All expected guarantees of the form $\mathbb{E}[\|\mathcal{G}_{F,\eta}(\overline{x})\|^2] \leq \varepsilon^2$ or $\mathbb{E}[\|\mathcal{G}_{F,\eta}(\overline{x})\|] \leq \varepsilon$ throughout this paper can be made into high-confidence bound by repeating the algorithm multiple times, each time estimating the value of $\|\mathcal{G}_{F,\eta}(\overline{x})\|$ using roughly $O(\frac{\mathcal{V}}{\varepsilon^2})$ stochastic gradient computations, and finally outputting the point $\overline{x}$ that leads to the smallest value $\|\mathcal{G}_{F,\eta}(\overline{x})\|$.

## 6.1  Proof of Theorem 3

Recall that

$$F^{(0)}(x) \stackrel{\text{def}}{=} F(x) \quad \text{and} \quad F^{(s)}(x) \stackrel{\text{def}}{=} F^{(s-1)}(x) + \frac{\sigma_s}{2}\|x - \widehat{x}_s\|^2 \quad \text{for } s = 1, 2, \ldots, S$$

Before proving Theorem 3, we state a few properties regarding the relationships between the objective-optimality of $\widehat{x}_s$ and point distances.

**Claim 6.2.** *Suppose for every $s = 1, \ldots, S$ the vector $\widehat{x}_s$ satisfies*

$$\mathbb{E}\big[F^{(s-1)}(\widehat{x}_s) - F^{(s-1)}(x^*_{s-1})\big] \leq \delta_s \quad \text{where} \quad x^*_{s-1} \in \arg\min_x\{F^{(s-1)}(x)\} \ , \tag{6.1}$$

*then,*

(a) *for every $s \geq 1$, $\mathbb{E}[\|\widehat{x}_s - x^*_{s-1}\|]^2 \leq \mathbb{E}[\|\widehat{x}_s - x^*_{s-1}\|^2] \leq \frac{2\delta_s}{\sigma_{s-1}}$,*

**Algorithm 3** $\text{SGD3}^{\text{sc}}(F, x_0, \sigma, L, T)$

**Input:** function $F(x) = \psi(x) + \frac{1}{n}\sum_{i=1}^{n} f_i(x)$; initial vector $x_0$; parameters $0 < \sigma \leq L$; number of iterations $T \geq \Omega\left(\frac{L}{\sigma}\log\frac{L}{\sigma}\right)$. $\diamond$ $f(x) = \frac{1}{n}\sum_{i=1}^{n} f_i(x)$ is $\sigma$-strongly convex and $L$-smooth

1: $F^{(0)}(x) \stackrel{\text{def}}{=} F(x)$; $\widehat{\mathsf{x}}_0 \leftarrow x_0$; $\sigma_0 \leftarrow \sigma$;
2: **for** $s = 1$ **to** $S = \lfloor\log_2\frac{L}{\sigma}\rfloor$ **do**
3: $\quad \widehat{\mathsf{x}}_s \leftarrow \text{SGD}^{\text{sc}}\left(F^{(s-1)}, \widehat{\mathsf{x}}_{s-1}, \sigma_{s-1}, 3L, \frac{T}{S}\right)$;
4: $\quad \sigma_s \leftarrow 2\sigma_{s-1}$;
5: $\quad F^{(s)}(x) \stackrel{\text{def}}{=} F^{(s-1)}(x) + \frac{\sigma_s}{2}\|x - \widehat{\mathsf{x}}_s\|^2$;
6: **return** $\overline{x} = \widehat{\mathsf{x}}_S$.

---

**Algorithm 4** $\text{SGD3}(F, x_0, \sigma, L, T)$

**Input:** function $F(x) = \psi(x) + \frac{1}{n}\sum_{i=1}^{n} f_i(x)$; initial vector $x_0$; parameters $L \geq \sigma > 0$; $T \geq 1$. $\diamond$ $f(x) = \frac{1}{n}\sum_{i=1}^{n} f_i(x)$ is convex and $L$-smooth

1: $G(x) \stackrel{\text{def}}{=} F(x) + \frac{\sigma}{2}\|x - x_0\|^2$;
2: **return** $\overline{x} \leftarrow \text{SGD3}^{\text{sc}}(G, x_0, \sigma, L + \sigma, T)$.

---

(b) for every $s \geq 1$, $\mathbb{E}[\|\widehat{\mathsf{x}}_s - x_s^*\|]^2 \leq \mathbb{E}[\|x_s^* - \widehat{\mathsf{x}}_s\|^2] \leq \frac{\delta_s}{\sigma_s}$; and

(c) if $\sigma_s = 2\sigma_{s-1}$ for all $s \geq 1$, then $\mathbb{E}\left[\sum_{s=1}^{S} \sigma_s \|x_S^* - \widehat{\mathsf{x}}_s\|\right] \leq 4\sum_{s=1}^{S}\sqrt{\delta_s \sigma_s}$.

*Proof of Claim 6.2.*

(a) $\mathbb{E}[\|\widehat{\mathsf{x}}_s - x_{s-1}^*\|]^2 \stackrel{①}{\leq} \mathbb{E}[\|\widehat{\mathsf{x}}_s - x_{s-1}^*\|^2] \stackrel{②}{\leq} \frac{2}{\sigma_{s-1}}\mathbb{E}[F^{(s-1)}(\widehat{\mathsf{x}}_s) - F^{(s-1)}(x_{s-1}^*)] \leq \frac{2\delta_s}{\sigma_{s-1}}$. Here, inequality ① is because $\mathbb{E}[X]^2 \leq \mathbb{E}[X^2]$, and inequality ② is due to the strong convexity of $F^{(s-1)}(x)$.

(b) We derive that

$$\sigma_s \|x_s^* - \widehat{\mathsf{x}}_s\|^2 \stackrel{①}{\leq} \frac{\sigma_s}{2}\|x_s^* - \widehat{\mathsf{x}}_s\|^2 + F^{(s)}(\widehat{\mathsf{x}}_s) - F^{(s)}(x_s^*) = F^{(s-1)}(\widehat{\mathsf{x}}_s) - F^{(s-1)}(x_s^*)$$
$$\stackrel{②}{\leq} F^{(s-1)}(\widehat{\mathsf{x}}_s) - F^{(s-1)}(x_{s-1}^*) .$$

Here, inequality ① is due to the strong convexity of $F^{(s)}(x)$, and inequality ② is because of the minimality of $x_{s-1}^*$. Taking expectation we have $\mathbb{E}[\|x_s^* - \widehat{\mathsf{x}}_s\|]^2 \leq \mathbb{E}[\|x_s^* - \widehat{\mathsf{x}}_s\|^2] \leq \frac{\delta_s}{\sigma_s}$.

(c) Define $P_t \stackrel{\text{def}}{=} \sum_{s=1}^{t} \sigma_s \|x_t^* - \widehat{\mathsf{x}}_s\|$ for each $t \geq 0, 1, \ldots, S$. Then by triangle inequality we have

$$P_s - P_{s-1} \leq \sigma_s \|x_s^* - \widehat{\mathsf{x}}_s\| + \left(\sum_{t=1}^{s-1}\sigma_t\right) \cdot \|x_s^* - x_{s-1}^*\|$$

Using the parameter choice of $\sigma_s = 2\sigma_{s-1}$, and plugging in Claim 6.2(a) and Claim 6.2(b), we have

$$\mathbb{E}[P_s - P_{s-1}] \leq \sqrt{\delta_s \sigma_s} + \sigma_s \cdot \mathbb{E}\left[\|x_s^* - \widehat{\mathsf{x}}_s\| + \|x_{s-1}^* - \widehat{\mathsf{x}}_s\|\right] \leq 4\sqrt{\delta_s \sigma_s} . \quad \square$$

*Proof of Theorem 3(a).* We first note that, when writing $f^{(s-1)}(x) = F^{(s-1)}(x) - \psi(x)$, each $f^{(s-1)}$ is at least $\sigma_{s-1}$-strongly convex and $L + \sum_{t=1}^{s-1}\sigma_t \leq 3L$ Lipschitz smooth. Therefore, applying Theorem 4.1(b), we have

$$\mathbb{E}[F^{(s-1)}(\widehat{\mathsf{x}}_s) - F^{(s-1)}(x_{s-1}^*)] \leq O\left(\frac{S\mathcal{V}}{\sigma_{s-1}T}\right) + \left(1 - \frac{\sigma_{s-1}}{3L}\right)^{\Omega(T/S)}\mathbb{E}[\sigma_{s-1}\|\widehat{\mathsf{x}}_{s-1} - x_{s-1}^*\|^2] .$$

If $s = 1$, this means (recalling $\widehat{x}_0 = x_0$ and $x_0^* = x^*$)
$$\mathbb{E}[F^{(0)}(\widehat{x}_s) - F^{(0)}(x^*)] \leq O\Big(\frac{S\mathcal{V}}{\sigma_0 T}\Big) + \Big(1 - \frac{\sigma_0}{L}\Big)^{\Omega(T/S)} \sigma_0 \|x_0 - x^*\|^2 \ .$$

If $s > 1$, this means
$$\mathbb{E}[F^{(s-1)}(\widehat{x}_s) - F^{(s-1)}(x_{s-1}^*)] \leq O\Big(\frac{S\mathcal{V}}{\sigma_{s-1} T}\Big) + \Big(1 - \frac{\sigma_{s-1}}{L}\Big)^{\Omega(T/S)} \mathbb{E}[F^{(s-2)}(\widehat{x}_{s-1}) - F^{(s-2)}(x_{s-2}^*)] \ .$$

Together, this means to satisfy (6.1), it suffices to choose $\delta_s$ so that
$$\delta_s = O\Big(\frac{S\mathcal{V}}{\sigma_s T}\Big) + \Big(1 - \frac{\sigma_0}{L}\Big)^{\Omega(sT/S)} \sigma_0 \|x_0 - x^*\|^2 \ .$$

Using Lemma 2.3 with $F^{(S-1)}$ and $y = x = \widehat{x}_S$, we have $\frac{\eta}{2} \|\mathcal{G}_{F^{(S-1)},\eta}(\widehat{x}_S)\|^2 \leq F^{(S-1)}(\widehat{x}_S) - F^{(S-1)}(\widehat{x}_S^+) \leq F^{(S-1)}(\widehat{x}_S) - F^{(S-1)}(x_{S-1}^*)$ and therefore
$$\mathbb{E}\big[\|\mathcal{G}_{F^{(S-1)},\eta}(\widehat{x}_S)\|\big]^2 \leq \mathbb{E}\big[\|\mathcal{G}_{F^{(S-1)},\eta}(\widehat{x}_S)\|^2\big] \leq \frac{2\delta_S}{\eta} = O(L\delta_S) \ .$$

Plugging this into Lemma 5.1 (with $G(x) = F^{(S-1)}(x)$) and Claim 6.2(c), we have
$$\mathbb{E}\big[\|\mathcal{G}_{F,\eta}(\widehat{x}_S)\|\big] \leq \mathbb{E}\Big[\sum_{s=1}^{S-1} \sigma_s \|x_{S-1}^* - \widehat{x}_s\| + 3\|\mathcal{G}_{F^{(S-1)},\eta}(\widehat{x}_S)\|\Big] \leq O\Big(\sum_{s=1}^{S-1} \sqrt{\delta_s \sigma_s} + \sqrt{L\delta_S}\Big)$$
$$= O\Big(\sum_{s=1}^S \sqrt{\delta_s \sigma_s}\Big) \leq O\Big(\frac{S^{3/2}\mathcal{V}^{1/2}}{T^{1/2}}\Big) + \Big(1 - \frac{\sigma_0}{L}\Big)^{\Omega(T/S)} \sigma_0 \|x_0 - x^*\| \ . \qquad \square$$

*Proof of Theorem 3(b).* Define $G(x) \stackrel{\text{def}}{=} F(x) + \frac{\sigma}{2}\|x - x_0\|^2$ and let $x_G^*$ be the (unique) minimizer of $G(\cdot)$. Note that $x_G^*$ may be different from $x^*$ which is a minimizer of $F(\cdot)$. Applying Theorem 3(a) on $G(x)$ and Lemma 5.1 with $S = 1$ and $\widehat{x}_1 = x_0$, we have
$$\mathbb{E}[\|\mathcal{G}_{F,\eta}(\overline{x})\|] \leq O\Big(\sigma\|x_0 - x_G^*\| + \frac{\sqrt{\mathcal{V}} \cdot \log^{3/2} \frac{L}{\sigma}}{\sqrt{T}}\Big) + \Big(1 - \frac{\sigma}{L}\Big)^{\Omega(T/\log(L/\sigma))} \sigma \|x_0 - x_G^*\|$$

Now, by definition $\frac{\sigma}{2}\|x^* - x_0\|^2 - \frac{\sigma}{2}\|x_G^* - x_0\|^2 = (G(x^*) - F(x^*)) + (F(x_G^*) - G(x_G^*)) \geq 0$ so we have $\|x_G^* - x_0\| \leq \|x^* - x_0\|$. This completes the proof. $\square$

# 7 Applications to Non-Convex Optimization

In this section, we extend our techniques to non-convex optimization, by designing SGD variants to find approximate stationary points (in Section 7.1) and approximate local minima (in Section 7.2).

## 7.1 Finding Approximate Stationary Points

Consider the following *non-convex* variant of Problem (3.1):
$$\min_{x \in \mathbb{R}^d} \big\{ F(x) = \psi(x) + f(x) \stackrel{\text{def}}{=} \psi(x) + \tfrac{1}{n} \sum_{i \in [n]} f_i(x) \big\} \ , \tag{7.1}$$

where instead of assuming $f(x)$ to be $L$-smooth and $\sigma$-strongly convex, we assume

- $f(x)$ is $L$-smooth and $\sigma$-*nonconvex* for some $\sigma \in [0, L]$, meaning that $\forall x, y \in \mathbb{R}^d$, it satisfies $f(y) \geq f(x) + \langle \nabla f(x), y - x \rangle - \frac{\sigma}{2}\|x - y\|^2$.

Our main goal is to design an algorithm to find approximate stationary points of $F(x)$, namely, a point $\overline{x}$ with $\|\mathcal{G}_{F,\eta}(\overline{x})\| \leq \varepsilon$. (We shall consider the more ambitious goal to find approximate local minima in Section 7.2.)

**Algorithm 5** SGD4$(F, x_0, \sigma, L, T_0, T)$

**Input:** function $F(x) = \psi(x) + \frac{1}{n}\sum_{i=1}^{n} f_i(x)$; initial vector $x_0$; parameters $L \geq \sigma > 0$; $T_0 \geq \Omega(\frac{L}{\sigma})$;
  $T \geq \max\{T_0, \Omega(\frac{L}{\sigma}\log\frac{L}{\sigma})\}$. ⋄ $f(x) = \frac{1}{n}\sum_{i=1}^{n} f_i(x)$ is $\sigma$-nonconvex and $L$-smooth

1: $\widehat{x}_0 \leftarrow x_0$; $S \leftarrow \lceil \frac{T}{T_0}\rceil$;
2: **for** $s \leftarrow 0$ **to** $S - 1$ **do**
3: $\quad G^{(s)}(x) \stackrel{\text{def}}{=} F(x) + \sigma\|x - \widehat{x}_s\|^2$;
4: $\quad \widehat{x}_{s+1} \leftarrow$ SGD$^{\text{sc}}(G^{(s)}, \widehat{x}_s, \sigma, L + 2\sigma, T_0)$;
5: $s \leftarrow$ a uniform random index in $\{0, 1, \ldots, S - 1\}$;
6: **return** $\overline{x} \leftarrow$ SGD3$^{\text{sc}}(G^{(s)}, \widehat{x}_s, \sigma, L + 2\sigma, T)$;

We propose a new method SGD4 (see Algorithm 5) which, starting from a vector $\widehat{x}_0 = x_0$, recursively minimizes a regularized function $G^{(s)}(x) \stackrel{\text{def}}{=} F(x) + \sigma\|x - \widehat{x}_s\|^2$. Since $G^{(s)}(x)$ is $\sigma$-strongly convex, we can apply SGD$^{\text{sc}}$ to minimize $G^{(s)}(x)$ in terms of decreasing its objective value. Let the resulting point be $\widehat{x}_{s+1}$ and SGD4 moves to the next iteration. In the end, SGD4 selects a random point $\widehat{x}_s$ uniformly at random, and applies SGD3$^{\text{sc}}$ to minimize $G^{(s)}(x)$ in terms of decreasing the gradient norm.

We have the following main theorem for SGD4:

**Theorem 4** (SGD4). *Let $x^* \in \arg\min_x\{F(x)\}$. To solve Problem (7.1) given any starting vector $x_0 \in \mathbb{R}^d$ and $\eta = \frac{C}{L}$ any absolute constant $C \in (0, \frac{1}{3}]$, suppose $T_0 \geq \Omega(\frac{L}{\sigma})$ and $T \geq \max\{T_0, \Omega(\frac{L}{\sigma}\log\frac{L}{\sigma})\}$, then SGD4$(F, x_0, \sigma, L, T_0, T)$ outputs $\overline{x}$ in gradient complexity $O(T)$ with*

$$\mathbb{E}[\|\mathcal{G}_{F,\eta}(\overline{x})\|] \leq O\Big(\frac{\sqrt{\sigma(F(x_0) - F(x^*))}}{\sqrt{T/T_0}} + \frac{\sqrt{\mathcal{V}}}{\sqrt{T_0}} + \frac{\sqrt{\mathcal{V}} \cdot \log^{3/2}\frac{L}{\sigma}}{\sqrt{T}}\Big) \ .$$

(For the same reason as Remark 6.1, the above expected guarantee can be made in high confidence.)

**Corollary 7.1.** *In other words, for every $T \geq \Omega(\frac{L}{\sigma}\log\frac{L}{\sigma})$, if $T_0$ is chosen optimally, $\overline{x}$ satisfies*

$$\mathbb{E}[\|\mathcal{G}_{F,\eta}(\overline{x})\|] \leq O\Big(\frac{\sqrt{L(F(x_0) - F(x^*))}}{\sqrt{T}} + \frac{(\mathcal{V}\sigma(F(x_0) - F(x^*)))^{1/4}}{T^{1/4}} + \frac{\sqrt{\mathcal{V}} \cdot \log^{3/2}\frac{L}{\sigma}}{\sqrt{T}}\Big) \ .$$

*Equivalently, to obtain a point $\overline{x}$ with $\mathbb{E}[\|\mathcal{G}_{F,\eta}(\overline{x})\|] \leq \varepsilon$, we need gradient complexity*

$$T = O\Big(\frac{L}{\sigma}\log\frac{L}{\sigma} + \frac{L(F(x_0) - F(x^*))}{\varepsilon^2} + \frac{\mathcal{V}\log^3(L/\sigma)}{\varepsilon^2} + \frac{\sigma\mathcal{V}(F(x_0) - F(x^*))}{\varepsilon^4}\Big) \ .$$

Before proving Theorem 4, we first state a simple variant of Lemma 5.1.

**Lemma 7.2** ([3]). *Suppose $\psi(x)$ is proper convex and $f(x)$ is $\sigma$-nonconvex and $L$-smooth. Consider $G(x) = F(x) + \sigma\|x - \widehat{x}\|$ which is $\sigma$-strongly convex. Let $x^*$ be the unique minimizer of $G(y)$, and $x$ be an arbitrary vector in the domain of $\{x \in \mathbb{R}^d : \psi(x) < +\infty\}$. Then,*

$$\forall \eta \in \big(0, \frac{1}{L + 2\sigma}\big] : \quad \|\mathcal{G}_{F,\eta}(x)\|^2 + \sigma^2\|x - \widehat{x}\|^2 \leq O\big(\sigma^2\|x^* - \widehat{x}\|^2 + \|\mathcal{G}_{G,\eta}(x)\|^2\big) \ .$$

(Lemma 7.2 appeared in [3, Lemma 3.5] and can be proved analogously to Lemma 5.1.)

*Proof of Theorem 4.* Define $G^{(s)}(x) \stackrel{\text{def}}{=} F(x) + \sigma\|x - \widehat{x}_s\|^2 = \psi(x) + f(x) + \sigma\|x - \widehat{x}_s\|^2$ and we have that $g^{(s)}(x) = f(x) + \sigma\|x - \widehat{x}_s\|^2$ is $\sigma$-strongly convex and $(L + 2\sigma)$-smooth. Let $x_s^*$ be the (unique) minimizer of $G^{(s)}(\cdot)$.

Since $\widehat{x}_{s+1} = \text{SGD}^{\text{sc}}(G^{(s)}, \widehat{x}_s, \sigma, L + 2\sigma, T_0)$, applying Theorem 4.1(b) we have

$$\mathbb{E}[G^{(s)}(\widehat{x}_{s+1})] - G^{(s)}(x_s^*) \leq O\left(\frac{\mathcal{V}}{\sigma T_0}\right) + \left(1 - \frac{\sigma}{L}\right)^{\Omega(T_0)} \sigma \|\widehat{x}_s - x_s^*\|^2 \ .$$

Using the identity formula $G^{(s)}(\widehat{x}_{s+1}) = G^{(s)}(\widehat{x}_s) - (F(\widehat{x}_s) - F(\widehat{x}_{s+1})) + \sigma \|\widehat{x}_{s+1} - \widehat{x}_s\|^2$, and the strong convexity which says $\frac{\sigma}{2} \|\widehat{x}_s - x_s^*\|^2 \leq G^{(s)}(\widehat{x}_s) - G^{(s)}(x_s^*)$, we have

$$\frac{\sigma}{2} \|\widehat{x}_s - x_s^*\|^2 + \mathbb{E}[\sigma \|\widehat{x}_{s+1} - \widehat{x}_s\|^2] \leq (F(\widehat{x}_s) - \mathbb{E}[F(\widehat{x}_{s+1})]) + O\left(\frac{\mathcal{V}}{\sigma T_0}\right) + \left(1 - \frac{\sigma}{L}\right)^{\Omega(T_0)} \sigma \|\widehat{x}_s - x_s^*\|^2 \ .$$

Since $T_0 \geq \Omega(L/\sigma)$, we can write

$$\mathbb{E}\left[\frac{\sigma}{4} \|\widehat{x}_s - x_s^*\|^2 + \sigma \|\widehat{x}_{s+1} - \widehat{x}_s\|^2\right] \leq (F(\widehat{x}_s) - \mathbb{E}[F(\widehat{x}_{s+1})]) + O\left(\frac{\mathcal{V}}{\sigma T_0}\right) \ . \tag{7.2}$$

After telescoping (7.2) for $s = 0, 1, \ldots, S-1$ and selecting $s \in \{0, 1, \ldots, S-1\}$ at random, we have

$$\mathbb{E}\left[\frac{\sigma}{4} \|\widehat{x}_s - x_s^*\|^2\right] \leq \frac{F(\widehat{x}_0) - F(x^*)}{S} + O\left(\frac{\mathcal{V}}{\sigma T_0}\right) \ . \tag{7.3}$$

For this particular choice of $s$, since $\overline{x} = \text{SGD3}^{\text{sc}}(G^{(s)}, \widehat{x}_s, \sigma, L + 2\sigma, T)$, Theorem 3(a) gives

$$\mathbb{E}[\|\mathcal{G}_{G^{(s)}, \eta}(\overline{x})\|] \leq O\left(\frac{\sqrt{\mathcal{V}} \cdot \log^{3/2} \frac{L}{\sigma}}{\sqrt{T}}\right) + \left(1 - \frac{\sigma}{L}\right)^{\Omega(T/\log(L/\sigma))} \sigma \|\widehat{x}_s - x_s^*\| \ .$$

Since $T \geq \Omega(\frac{L}{\sigma} \log \frac{L}{\sigma})$, this implies

$$\mathbb{E}[\|\mathcal{G}_{G^{(s)}, \eta}(\overline{x})\|] \leq O\left(\frac{\sqrt{\mathcal{V}} \cdot \log^{3/2} \frac{L}{\sigma}}{\sqrt{T}}\right) + \frac{\sigma}{2} \|\widehat{x}_s - x_s^*\| \ . \tag{7.4}$$

Applying Lemma 7.2 and $\sqrt{a^2 + b^2} \leq (a+b)$, we have

$$\|\mathcal{G}_{F, \eta}(\overline{x})\| \leq O\left(\sigma \|\widehat{x}_s - x_s^*\| + \|\mathcal{G}_{G^{(s)}, \eta}(\overline{x})\|\right) \ . \tag{7.5}$$

Finally, combining (7.3), (7.4) and (7.5), and the fact that $\mathbb{E}[\|\cdot\|]^2 \leq \mathbb{E}[\|\cdot\|^2]$, we have

$$\mathbb{E}[\|\mathcal{G}_{F,\eta}(\overline{x})\|] \leq O\left(\frac{\sqrt{\sigma(F(\widehat{x}_0) - F(x^*))}}{\sqrt{S}} + \frac{\sqrt{\mathcal{V}}}{\sqrt{T_0}} + \frac{\sqrt{\mathcal{V}} \cdot \log^{3/2} \frac{L}{\sigma}}{\sqrt{T}}\right) \ .$$

This completes the proof using $S = \lceil \frac{T}{T_0} \rceil$ and $\widehat{x}_0 = x_0$. $\square$

## 7.2 Finding Approximate Local Minima

Consider the following *non-convex* variant of Problem (3.1):

$$\min_{x \in \mathbb{R}^d} \left\{ f(x) \stackrel{\text{def}}{=} \frac{1}{n} \sum_{i \in [n]} f_i(x) \right\} \ , \tag{7.6}$$

where[9]

1. each $f_i(x)$ is possibly nonconvex but $L$-smooth,

2. the average $f(x)$ is possibly nonconvex, but $L_2$-second-order smooth, and

3. the stochastic gradients $\nabla f_i(x)$ have a bounded variance, that is

$$\forall x \in \mathbb{R}^d: \quad \mathbb{E}_{i \in_R [n]} \|\nabla f(x) - \nabla f_i(x)\|^2 \leq \mathcal{V} \ .$$

---

[9]Like in previous works, we do not include the proximal term $\psi(\cdot)$ when finding approximate local minima, because it can be tricky to define what local minima mean when $\psi(\cdot)$ is present. Also, recall that second-order smoothness is a necessary condition for finding approximate local minima.

**Algorithm 6** `SGD5`$(f, y_0, \varepsilon, \delta)$

**Input:** function $f(x)$ satisfying Problem (7.6), starting vector $y_0$, target accuracy $\varepsilon > 0$ and $\delta > 0$.

⋄ assume $\mathcal{V} \geq \Omega(\varepsilon^2)$ and $\delta \leq O(\sqrt{\varepsilon L_2})$

1: **if** $L_2 \geq \frac{L\delta}{\varepsilon}$ **then** $\widetilde{L} = \widetilde{\sigma} \leftarrow \Theta(\frac{\varepsilon L_2}{\delta}) \in [L, \infty)$.      ⋄ the boundary case for large $L_2$
2: **else** $\widetilde{\sigma} \leftarrow \Theta\big(\max\big\{\frac{\varepsilon L_2}{\delta}, \frac{\varepsilon^2 L}{\mathcal{V}}\big\}\big) \in [\delta, L]$ and $\widetilde{L} \leftarrow L$.      ⋄ the interesting case
3: $X \leftarrow [\,]$.
4: $B \leftarrow \Theta(\mathcal{V}/\varepsilon^2)$ and $N_1 \leftarrow \Theta\big(\frac{\widetilde{\sigma}\Delta_f}{\varepsilon^2}\big)$, where $\Delta_f$ is an upper bound on $f(y_0) - \min_y\{f(y)\}$.
5: **for** $k \leftarrow 0$ **to** $\infty$ **do**
6:      Apply Oja's algorithm to find minEV of $\nabla^2 f(y_k)$.    ⋄ use Lemma E.3 with $T_{\mathsf{oja}} = \Theta(\frac{L^2}{\delta^2}\log(dk))$
7:      **if** $v \in \mathbb{R}^d$ is found s.t. $v^\top \nabla^2 f(y_k) v \leq -\frac{\delta}{2}$ **then**
8:          $y_{k+1} \leftarrow y_k \pm \frac{\delta}{L_2} v$ where the sign is random.
9:      **else**      ⋄ it satisfies $\nabla^2 f(y_k) \succeq -\delta \mathbf{I}$ w.p. $\geq 1 - \frac{1}{20(k+1)^2}$, see Lemma E.3.
10:          $F(x) = F^k(x) \stackrel{\mathrm{def}}{=} f(x) + L(\max\{0, \|x - y_k\| - \frac{\delta}{L_2}\})^2$.
     ⋄ $F(\cdot)$ is $3L$-smooth and $5\sigma$-nonconvex
11:          $G(x) = G^k(x) \stackrel{\mathrm{def}}{=} F^k(x) + \widetilde{\sigma}\|x - y_k\|^2$;      ⋄ $G(\cdot)$ is $O(\widetilde{L})$-smooth and $\widetilde{\sigma}$-strongly convex
12:          $y_{k+1} \leftarrow \mathtt{SGD^{sc}}(G, y_k, O(\widetilde{\sigma}), O(\widetilde{L}), B)$;
13:          $X \leftarrow [X, y_k]$.
14:      break the for loop if have performed $N_1$ first-order steps.
15: $y \leftarrow$ a random vector in $X$.
16: define $G(x) \stackrel{\mathrm{def}}{=} f(x) + L(\max\{0, \|x - y\| - \frac{\delta}{L_2}\})^2 + \widetilde{\sigma}\|x - y\|^2$.      ⋄ $G(x)$ is $\widetilde{\sigma}$-strongly convex
17: $x^{\mathsf{out}} \leftarrow \mathtt{SGD3^{sc}}(G, y, \widetilde{\sigma}, O(\widetilde{L}), T_{\mathsf{sgd}})$.      ⋄ $T_{\mathsf{sgd}} = \Theta\big(\frac{\mathcal{V}}{\varepsilon^2}\log^3 \frac{\widetilde{L}}{\widetilde{\sigma}}\big)$
18: **return** $x^{\mathsf{out}}$.

---

Our goal here is to find an $(\varepsilon, \delta)$-approximate local minimum of $f(x)$, that is, a point $x$ satisfying $\|\nabla f(x^{\mathsf{out}})\| \leq \varepsilon$ and $\nabla^2 f(x^{\mathsf{out}}) \succeq -3\delta \mathbf{I}$.

We propose `SGD5` (see Algorithm 6) to solve this task, and `SGD5` follows the exact same "swing by saddle point" framework of [3].[10] `SGD5` starts from a vector $y_0 \in \mathbb{R}^d$ and is divided into iterations $k = 0, 1, \ldots$. In each iteration $k$, it *either* finds a vector $v \in \mathbb{R}^d$ such that $v^\top \nabla^2 f(y_k) v \leq -\frac{\delta}{2}$, *or* conclude that $\nabla^2 f(y_k) \succeq -\delta I$. This can be done by running Oja's algorithm of Allen-Zhu and Li [5] for $\widetilde{O}(L^2/\delta^2)$ iterations (see Section E.1 for completeness' sake).

- If $v^\top \nabla^2 f(y_k) v \leq -\frac{\delta}{2}$, we choose $y_{k+1} \leftarrow y_k + \frac{\delta}{L_2} v$ and $y_{k+1} \leftarrow y_k - \frac{\delta}{L_2} v$ each with probability $1/2$. We call this a second-order step.

- If $\nabla^2 f(y_k) \succeq -\delta I$, then we define $G(x) = G^k(x) \stackrel{\mathrm{def}}{=} f(x) + L(\max\{0, \|x - y_k\| - \frac{\delta}{L_2}\})^2 + \widetilde{\sigma}\|x - y_k\|^2$, and apply `SGD3`$^{\mathsf{sc}}$ for $O(\mathcal{V}/\varepsilon^2)$ iterations to minimize $G(x)$. We call this a first-order step, and move to a new point $y_{k+1}$ which is the output of `SGD3`$^{\mathsf{sc}}$. (In Line 2 of `SGD5`, we choose $\widetilde{\sigma} \geq \delta$ carefully in order to deal with several different parameter regimes.)

In the end, we terminate `SGD5` whenever $N_1$ iterations of first-order steps are executed. We select a random $y$ along the $N_1$ first-order steps, and find a point $x^{\mathsf{out}}$ which gives small gradient for $G(x)$ using `SGD`$^{\mathsf{sc}}$.

We state the main theorem of `SGD5` as follows. Its proof is a simple combination of the proof of Theorem 4 and the "swing by saddle point" technique of [3]. We include them in Appendix E only for completeness' sake.

---

[10] The only algorithmic change is to replace the use of `Natasha1.5` method of [3] by simpler SGD variants.

> **Theorem 5** (SGD5). *Consider Problem (7.6) with a starting vector $y_0$. For any $\varepsilon > 0$ and $\delta \in (0, L]$, under assumptions $\mathcal{V} \geq \Omega(\varepsilon^2)$ and $\delta \leq O(\sqrt{\varepsilon L_2})$, the output $x^{\text{out}} = \text{SGD5}(f, y_0, \varepsilon, \delta)$ satisfies, with probability at least $2/3$,*
> $$\|\nabla f(x^{\text{out}})\| \leq \varepsilon \quad \text{and} \quad \nabla^2 f(x^{\text{out}}) \succeq -3\delta \mathbf{I} \ .$$
> *The total gradient complexity $T$ is*
> $$T = \widetilde{O}\Big(\frac{\mathcal{V}}{\varepsilon^2} + \frac{L_2^2 \Delta_f}{\delta^3} \cdot \frac{L^2}{\delta^2} + \frac{L_2 \Delta_f}{\varepsilon \delta} \cdot \frac{\mathcal{V}}{\varepsilon^2} + \frac{L \Delta_f}{\mathcal{V}} \cdot \frac{L^2}{\delta^2}\Big) \ .$$
> *Above, $\Delta_f$ is any known upper bound on $f(y_0) - \min_y \{f(y)\}$.*

*Remark* 7.3. In practice, one can just choose $N_1$, the number of first-order updates in SGD5, as sufficiently large, without the necessity of knowing $\Delta_f$.

**Corollary 7.4.** *If we assume $L, L_2, \Delta_f$ and $\mathcal{V}$ are constants, then SGD5 finds $x^{\text{out}}$ satisfying*
$$\|\nabla f(x^{\text{out}})\| \leq \varepsilon \quad \text{and} \quad \nabla^2 f(x^{\text{out}}) \succeq -\delta \mathbf{I}$$
*in gradient complexity $T = \widetilde{O}\big(\frac{1}{\delta^5} + \frac{1}{\delta \varepsilon^3}\big)$ for $\delta \in (0, \sqrt{\varepsilon}]$. Since when $\delta > \varepsilon^{1/2}$, we can set $\delta = \varepsilon^{1/2}$, this can be re-written as $T = \widetilde{O}\big(\frac{1}{\delta^5} + \frac{1}{\varepsilon^{3.5}} + \frac{1}{\delta \varepsilon^3}\big)$.*

## Acknowledgements

We would like to thank Lin Xiao for suggesting reference [31, Lemma 3.7], an anonymous researcher from the Simons Institute for suggesting reference [26], Yurii Nesterov for helpful discussions, Xinyu Weng for discussing the motivations, Sébastien Bubeck, Yuval Peres, and Lin Xiao for discussing notations, Chi Jin for discussing reference [29], and Dmitriy Drusvyatskiy for discussing the notion of Moreau envelope.

# Appendix

## A  Other Related Work

**Offline Convex Stochastic Optimization.** One can also ask the question of finding a point with small gradient when the convex function $f(x) = \frac{1}{n} \sum_{i=1}^n f_i(x)$ is a *finite* sum of functions. This is the *finite-sum stochastic* or *offline stochastic* setting. In this setting, the number of stochastic gradient computations —which we denote as the gradient complexity $T$— can depend on $n$.

For instance, if each $f_i(x)$ is $L$-smooth and $f(x)$ is $\sigma$-strongly convex, then the gradient complexity $T$ can be made $T \propto O\big((n + \sqrt{nL/\sigma}) \cdot \log \frac{1}{\delta}\big)$ to achieve a point with $f(x) - f(x^*) \leq \delta$, see for instance the Katyusha method [2] and the references therein. Since this is a linear-convergence rate, it translates to $T \propto O\big((n + \sqrt{nL/\sigma}) \cdot \log \frac{L}{\varepsilon}\big)$ for finding a point with $\|\nabla f(x)\| \leq \varepsilon$.

If $f(x)$ is convex but not strongly convex, then one can apply Nesterov's second trick to regularize $f(x)$ and then apply Katyusha. This gives gradient complexity $T \propto O\big((n + \sqrt{nL/\varepsilon}) \cdot \log \frac{L}{\varepsilon}\big)$.

In both cases, the offline stochastic method is no slower than the full-gradient ones (such as AGD), and known to be optimal [30]. We summarize them in Table 2 for comparison purpose only.

**Offline Non-Convex Stochastic Optimization.** One can similarly ask the questions of finding approximate stationary points, or approximate local minima, for a nonconvex function $f(x) =$ $\frac{1}{n}\sum_{i=1}^{n} f_i(x)$ that is a *finite* sum of $f_i(x)$. There is a lot of recent progress for these two problems, and we refer interested readers to the cited references in [3].

|  | **algorithm** | **gradient complexity $T$** | |
|---|---|---|---|
| offline convex | gradient descent (GD) | $O(n\varepsilon^{-2})$ | (see [26]) |
|  | accelerated gradient descent (AGD) | $O(n\varepsilon^{-1})$ | (see [26]) |
|  | GD after GD | $O(n\varepsilon^{-1})$ | (see [26]) |
|  | GD after AGD | $O(n\varepsilon^{-2/3})$ | (see [26]) |
|  | AGD + regularization | $O(n\varepsilon^{-1/2}\log\frac{1}{\varepsilon})$ | (see [26]) |
|  | Katyusha + regularization | $O\big((n+\sqrt{n}\cdot\varepsilon^{-1/2})\cdot\log\frac{1}{\varepsilon}\big)$ | [2] + [26] |
| offline strongly convex | GD | $O(n\cdot\kappa\cdot\log\frac{1}{\varepsilon})$ | (see [24]) |
|  | AGD | $O(n\cdot\sqrt{\kappa}\cdot\log\frac{1}{\varepsilon})$ | (see [24]) |
|  | Katyusha | $O\big((n+\sqrt{n\kappa})\cdot\log\frac{1}{\varepsilon}\big)$ | (see [2]) |

Table 2: Comparison of first-order **offline** methods for finding $\|\nabla f(x)\| \leq \varepsilon$. This table is for **reference only**. Following tradition, in these complexity bounds, we assume the smoothness parameters as constants, and only show the dependency on $n, \varepsilon$ and the condition number $\kappa = \frac{L}{\sigma} \geq 1$ (if the objective is strongly convex).

**Graduated Regularization.** Of course, the idea of gradually changing the parameter $\sigma$ for the weight of the regularizer is not totally new. For instance, when reducing weakly-convex optimization to strongly-convex optimization (both in terms of convergence in objective value), one can keep halving the value of $\sigma$ for a logarithmic number of rounds [4]. In contrast, we are doubly the value of $\sigma$ in SGD3. To the best of our knowledge, the analysis in [4] (and the references therein) cannot be applied to this paper.

**Non-Smooth Objectives.** Finding approximate stationary points may not be possible without any smoothness assumption on $f(x)$.[11] Thus, what can we do for non-smooth functions? At least for functions with $\sigma$-bounded nonconvexity, there is a meaningful alternative notion: namely, to minimize the gradient of the so-called Moreau envelope: $F_\lambda(x) \stackrel{\text{def}}{=} \min_y\{F(y) + \frac{\lambda}{2}\|y-x\|^2\}$ for any $\lambda > 2\sigma$. $F_\lambda(x)$ is well-defined and smooth because $F(y) + \frac{\lambda}{2}\|y-x\|^2$ is strongly convex in $y$. It can be shown (in a similar way as Lemma 7.2) that a point $x$ with small gradient for $F_\lambda(x)$ must be "close" to a point $\widehat{x}$ with small (sub-)gradient for $F(\widehat{x})$. Although finding $\widehat{x}$ may be computationally intractable, one can apply smooth methods to find $x$ with small gradients for Moreau envelope. We refer interested readers to [12, 13] and the references therein.

# B Approach 1: SGD After SGD

In this section, we generalize Nesterov's first trick to the stochastic setting. Namely, instead of directly turning a point $\overline{x}$ with good objective value into one with small gradient using (4.1), we wish to apply multiple steps of SGD to prune it.

More specifically, recall in Nesterov's first trick, he started from $\overline{x}$ and applied $T$ steps of GD for pruning. This implies $\frac{\eta}{2}\sum_{t=1}^{T}\|\mathcal{G}_{F,\eta}(x_t)\|^2 \leq F(\overline{x}) - F(x^*)$, and thus gives a point $x$ that has a gradient $T$ times smaller than before; in contrast, in (4.1) we only had $\frac{\eta}{2}\|\mathcal{G}_{F,\eta}(\overline{x})\|^2 \leq F(\overline{x}) - F(x^*)$.

In our stochastic setting, we start from $\overline{x}$ that is calculated from either `SGD` or `SGD`[sc]. Then, we

---
[11]For instance, if $f(x) = |x|$, then finding a point with $|\nabla f(x)| \leq 0.5$ would mean exactly finding $x^*$.

**Algorithm 7** $\texttt{SGD1}(F, x_0, \alpha, T, \eta, T_1)$

**Input:** function $F(x) = \psi(x) + \frac{1}{n}\sum_{i=1}^n f_i(x)$; initial vector $x_0$; learning rate $\alpha > 0$; $T \geq 1$; $T_1 \in [T]$.
1: $x_1 \leftarrow \texttt{SGD}(F, x_0, \alpha, T)$;
2: **for** $t = 1$ **to** $T_1$ **do**
3:     $S \leftarrow$ a uniform random subset of $[n]$ with cardinality $T/T_1$;
4:     $x_{t+1} \leftarrow \arg\min_{y \in \mathbb{R}^d}\{\psi(y) + \frac{1}{2\eta}\|y - x_t\|^2 + \frac{1}{|S|}\sum_{i \in S}\langle \nabla f_i(x_t), y\rangle\}$;
5: **return** $\overline{x} = x_t$ where $t \in [T_1]$ is uniformly chosen at random.

---

**Algorithm 8** $\texttt{SGD1}^{\mathsf{sc}}(F, x_0, \sigma, L, T, \eta, T_1)$

**Input:** function $F(x) = \psi(x) + \frac{1}{n}\sum_{i=1}^n f_i(x)$; vector $x_0$; parameters $0 < \sigma \leq L$; $T \geq \frac{L}{\sigma}$; $T_1 \in [T]$.
1: $x_1 \leftarrow \texttt{SGD}^{\mathsf{sc}}(F, x_0, \sigma, L, T)$;
2: **for** $t = 1$ **to** $T_1$ **do**
3:     $S \leftarrow$ a uniform random subset of $[n]$ with cardinality $T/T_1$;
4:     $x_{t+1} \leftarrow \arg\min_{y \in \mathbb{R}^d}\{\psi(y) + \frac{1}{2\eta}\|y - x_t\|^2 + \frac{1}{|S|}\sum_{i \in S}\langle \nabla f_i(x_t), y\rangle\}$;
5: **return** $\overline{x} = x_t$ where $t \in [T_1]$ is uniformly chosen at random.

---

apply $T_1$ steps of SGD, each with mini-batch size $b = T/T_1$. We show that it satisfies

$$\tfrac{\eta}{8}\mathbb{E}\big[\textstyle\sum_{t=1}^{T_1}\|\mathcal{G}_{F,\eta}(x_t)\|^2\big] - \tfrac{12\eta\mathcal{V}}{T/T_1} \leq F(\overline{x}) - F(x^*) \enspace,$$

and use this to replace the use of inequality (4.1). We summarize the resulting algorithms as $\texttt{SGD1}$ and $\texttt{SGD1}^{\mathsf{sc}}$, and prove the following theorem (see subsequent subsections):

> **Theorem 1** (SGD1). *Let $x^* \in \arg\min_x\{F(x)\}$. To solve Problem (3.1) given a starting vector $x_0 \in \mathbb{R}^d$ and $\eta = \frac{C}{L} \leq \frac{1}{4L}$ for some constant $C \leq 1/4$,*
>
> *(a) If $\alpha$ and $T_1$ are appropriately chosen, then $\texttt{SGD1}(F, x_0, \alpha, T, \eta, T_1)$ outputs $\overline{x}$ satisfying*
>
> $$\mathbb{E}[\|\mathcal{G}_{F,\eta}(\overline{x})\|^2] \leq O\Big(\tfrac{L^2\|x_0 - x^*\|^2}{T^2} + \tfrac{L\sqrt{\mathcal{V}}\|x_0 - x^*\|}{T} + \tfrac{\mathcal{V}}{T} + \tfrac{L^{1/2}\mathcal{V}^{3/4}\|x_0 - x^*\|^{1/2}}{T^{3/4}}\Big) \enspace.$$
>
> *(b) If $f(x)$ is $\sigma$-strongly convex for $\sigma \in (0, L]$, $T \geq \frac{L}{\sigma}$, and $T_1$ is appropriately chosen, then $\texttt{SGD1}^{\mathsf{sc}}(F, x_0, \sigma, L, T, \eta, T_1)$ outputs $\overline{x}$ satisfying*
>
> $$\mathbb{E}[\|\mathcal{G}_{F,\eta}(\overline{x})\|^2] \leq O\Big(\tfrac{\sqrt{L}\mathcal{V}}{\sqrt{\sigma}T}\Big) + \big(1 - \tfrac{\sigma}{L}\big)^{\Omega(T)} L\sigma\|x_0 - x^*\|^2 \enspace.$$

(For the same reason as Remark 6.1, the above expected guarantee can be made in high confidence.)

In the special case $\psi(x) \equiv 0$, Theorem 1(a) is simple to prove among experts. For instance, when $\psi(x) \equiv 0$, this $T \propto \varepsilon^{-8/3}$ rate was recorded by Ghadimi and Lan [17] using a more involved algorithm.[12] We are not aware of Theorem 1(b) being recorded before.

### B.1 Proof of Theorem 1(a)

The following fact says the variance of a random variable decreases by a factor $m$ if we choose $m$ independent copies and average them. It is trivial to prove.

---

[12]Ghadimi and Lan [17] showed this $T \propto \varepsilon^{-8/3}$ rate using an accelerated version of SGD. Note that acceleration only helps in reducing lower-order terms in the convergence rate, but is unnecessary for achieving $T \propto \varepsilon^{-8/3}$.

**Fact B.1.** If $v_1, \ldots, v_n \in \mathbb{R}^d$ satisfy $\sum_{i=1}^n v_i = \vec{0}$, and $S$ is a non-empty, uniform random subset of $[n]$. Then
$$\mathbb{E}\left[\left\|\tfrac{1}{|S|}\sum_{i\in S} v_i\right\|^2\right] = \tfrac{n-|S|}{(n-1)|S|} \cdot \tfrac{1}{n}\sum_{i\in[n]} \|v_i\|^2 \leq \tfrac{\mathbb{I}[|S|<n]}{|S|} \cdot \tfrac{1}{n}\sum_{i\in[n]} \|v_i\|^2 \ .$$

*Proof of Theorem 1(a).* We first apply Theorem 4.1(a) and obtain a point $x_1$ satisfying $\mathbb{E}[F(x_1)] - F(x^*) \leq O\left(\frac{L\|x_0-x^*\|^2}{T} + \frac{\sqrt{\mathcal{V}}\|x_0-x^*\|}{\sqrt{T}}\right)$, with total gradient complexity $T$.

Next, we start from $x_1$ and perform $T_1$ iterations of SGD, each time with mini-batch size $T/T_1$: that is, in each iteration $t = 1, \ldots, T_1$, we update
$$x_{t+1} = \arg\min_{y \in \mathbb{R}^d}\{\psi(y) + \frac{1}{2\eta}\|y - x_t\|^2 + \langle \nabla f_S(x_t), y\rangle\}$$

where $f_S(x) \stackrel{\mathrm{def}}{=} \frac{1}{|S|}\sum_{i \in S} f_i(x)$ and $S$ is a uniform random subset of $[n]$ for each iteration $t$, with cardinality $|S| = T/T_1$. Note that $T_1$ steps of mini-batch SGD only requires gradient complexity $T_1 \cdot \frac{T}{T_1} = T$. We wish to show that, focusing on one iteration from $x_t$ to $x_{t+1}$, we have
$$F(x_t) - \mathbb{E}_S[F(x_{t+1})] \geq \frac{\eta}{8}\mathbb{E}_S\left[\|\mathcal{G}_{F,\eta}(x_t)\|^2\right] - \frac{12\eta\mathcal{V}}{|S|} \ . \tag{B.1}$$

To prove (B.1), we denote by $x = x_t$ and by
$$z = \arg\min_y\left\{\psi(y) + \langle\nabla f(x), y\rangle + \frac{1}{2\eta}\|y-x\|^2\right\} = \arg\min_y\left\{\psi(y) + \langle\nabla f(x) - \frac{x}{\eta}, y\rangle + \frac{1}{2\eta}\|y\|^2\right\}$$
$$z_S = \arg\min_y\left\{\psi(y) + \langle\nabla f_S(x), y\rangle + \frac{1}{2\eta}\|y-x\|^2\right\} = \arg\min_y\left\{\psi(y) + \langle\nabla f_S(x) - \frac{x}{\eta}, y\rangle + \frac{1}{2\eta}\|y\|^2\right\}$$

We have by definition $\mathcal{G}_{F,\eta}(x) = \frac{1}{\eta}(x-z)$ and $z_S = x_{t+1}$.

For analysis purpose, let $g(y) \stackrel{\mathrm{def}}{=} \frac{1}{2\eta}\|y\|^2 + \psi(y)$ and recall the definition of Fenchel dual $g^*(\beta) = \max_y\{y^\top \beta - g(y)\}$. Proposition 2.5 says $\nabla g^*(\beta) = \max_y\{y^\top \beta - g(y)\}$. This implies $z = \nabla g^*(\frac{x}{\eta} - \nabla f(x))$ and $z_S = \nabla g^*(\frac{x}{\eta} - \nabla f_S(x))$. Therefore, using the property that $g^*(\cdot)$ is $\eta$-smooth (because $g(y)$ is $1/\eta$-strongly convex, see Proposition 2.6), we have
$$\|z - z_S\| \leq \eta\|\nabla f(x) - \nabla f_S(x)\| \ . \tag{B.2}$$

Next, we derive that
$$F(x) - F(z_S) = f(x) - f(z_S) + \psi(x) - \psi(z_S)$$
$$\stackrel{①}{\geq} \langle\nabla f(x), x - z_S\rangle - \frac{L}{2}\|x - z_S\|^2 + \psi(x) - \psi(z_S)$$
$$= \langle\nabla f(x) - \nabla f_S(x), x - z_S\rangle + \langle\nabla f_S(x), x - z_S\rangle - \frac{L}{2}\|x - z_S\|^2 + \psi(x) - \psi(z_S)$$
$$\stackrel{②}{\geq} \langle\nabla f(x) - \nabla f_S(x), x - z_S\rangle + \frac{1}{2\eta}\|x - z_S\|^2 - \frac{L}{2}\|x - z_S\|^2$$
$$\stackrel{③}{\geq} -2\eta\|\nabla f(x) - \nabla f_S(x)\|^2 - \frac{1}{8\eta}\|x - z_S\|^2 + \frac{1}{2\eta}\|x - z_S\|^2 - \frac{L}{2}\|x - z_S\|^2$$
$$\stackrel{④}{\geq} \frac{1}{4\eta}\|x - z_S\|^2 - 2\eta\|\nabla f(x) - \nabla f_S(x)\|^2$$
$$\stackrel{⑤}{\geq} \frac{1}{8\eta}\|x - z\|^2 - \frac{1}{4\eta}\|z - z_S\|^2 - 2\eta\|\nabla f(x) - \nabla f_S(x)\|^2$$
$$\stackrel{⑥}{\geq} \frac{1}{8\eta}\|x - z\|^2 - \frac{9}{4}\eta\|\nabla f(x) - \nabla f_S(x)\|^2$$

$$= \frac{\eta}{8}\|\mathcal{G}_{F,\eta}(x)\|^2 - \frac{9}{4}\eta\|\nabla f(x) - \nabla f_S(x)\|^2 \ . \tag{B.3}$$

Above, ① uses the smoothness of $f(\cdot)$; ② uses the definition of $z_S$ which implies $\psi(z_S)+\langle\nabla f_S(x), z_S\rangle + \frac{1}{2\eta}\|z_S - x\|^2 \leq \psi(x) + \langle\nabla f_S(x), x\rangle$; ③ uses Young's inequality; ④ uses $\eta \leq \frac{1}{4L}$; ⑤ uses AM-GM; ⑥ uses (B.2).

Next, we apply Fact B.1 (by letting $v_i = \nabla f(x) - \nabla f_i(x)$) and derive

$$\mathbb{E}_S\big[\|\nabla f(x) - \nabla f_S(x)\|^2\big] = \mathbb{E}\Big[\Big\|\frac{1}{|S|}\sum_{i\in S} v_i\Big\|^2\Big] \leq \frac{1}{|S|} \cdot \frac{1}{n}\sum_{i\in[n]}\|v_i\|^2 \leq \frac{\mathcal{V}}{|S|} \ ,$$

where the last inequality uses our assumption $\mathbb{E}_i\|\nabla f(x) - \nabla f_i(x)\|^2 \leq \mathcal{V}$. Plugging this back to (B.3), we finish the proof of (B.1).

Finally, we telescope (B.1) for all $t = 0, 1, \ldots, T_1 - 1$ and use $\eta = \Theta(1/L)$ to derive that

$$\clubsuit \stackrel{\text{def}}{=} \mathbb{E}\Big[\frac{1}{T_1}\sum_{t\in[T_1]}\|\mathcal{G}_{F,\eta}(x_t)\|^2\Big] \leq O\Big(\frac{L(F(x_1) - F(x^*))}{T_1} + \frac{\mathcal{V}}{|S|}\Big) \leq O\Big(\frac{L^2\|x_0 - x^*\|^2}{T_1 T} + \frac{L\sqrt{\mathcal{V}}\|x_0 - x^*\|}{T_1\sqrt{T}} + \frac{\mathcal{V}T_1}{T}\Big) \ .$$

There are two cases: $L\|x_0 - x^*\| \geq \sqrt{\mathcal{V}T}$ and $L\|x_0 - x^*\| \leq \sqrt{\mathcal{V}T}$.

- In the former case, we have $\clubsuit \leq O\Big(\frac{L^2\|x_0 - x^*\|^2}{T_1 T} + \frac{\mathcal{V}T_1}{T}\Big)$. After choosing $T_1 \in [1, T]$ to balance the two terms, we have $\clubsuit \leq O\Big(\frac{L^2\|x_0 - x^*\|^2}{T^2} + \frac{\mathcal{V}}{T} + \frac{L\sqrt{\mathcal{V}}\|x_0 - x^*\|}{T}\Big)$. It is easy to verify that the first term is always greater than the second. Therefore, $\clubsuit \leq O\Big(\frac{L^2\|x_0 - x^*\|^2}{T^2} + \frac{L\sqrt{\mathcal{V}}\|x_0 - x^*\|}{T}\Big)$.

- In the latter case, we have $\clubsuit \leq O\Big(\frac{L\sqrt{\mathcal{V}}\|x_0 - x^*\|}{T_1\sqrt{T}} + \frac{\mathcal{V}T_1}{T}\Big)$. After choosing $T_1 \in [1, T]$ to balance the two terms, we have $\clubsuit \leq O\Big(\frac{L\sqrt{\mathcal{V}}\|x_0 - x^*\|}{T^{3/2}} + \frac{\mathcal{V}}{T} + \frac{L^{1/2}\mathcal{V}^{3/4}\|x_0 - x^*\|^{1/2}}{T^{3/4}}\Big)$. The first term is always less than the second, so $\clubsuit \leq O\Big(\frac{\mathcal{V}}{T} + \frac{L^{1/2}\mathcal{V}^{3/4}\|x_0 - x^*\|^{1/2}}{T^{3/4}}\Big)$

In sum, we conclude $\clubsuit \leq O\Big(\frac{L^2\|x_0 - x^*\|^2}{T^2} + \frac{L\sqrt{\mathcal{V}}\|x_0 - x^*\|}{T} + \frac{\mathcal{V}}{T} + \frac{L^{1/2}\mathcal{V}^{3/4}\|x_0 - x^*\|^{1/2}}{T^{3/4}}\Big)$ so if we randomly output $x_1, \ldots, x_{T_1}$, we have the desired bound. □

### B.2 Proof of Theorem 1(b)

*Proof of Theorem 1(b).* We use the same proof of Theorem 1(a), except that we use $F(x_1) - F(x^*) \leq O\big(\frac{\mathcal{V}}{\sigma T}\big) + \big(1 - \frac{\sigma}{L}\big)^{\Omega(T)}\sigma\|x_0 - x^*\|^2$ from Theorem 4.1(b) instead of Theorem 4.1(a). Therefore, we have

$$\mathbb{E}\Big[\frac{1}{T_1}\sum_{t\in[T_1]}\|\mathcal{G}_{F,\eta}(x_t)\|^2\Big] \leq O\Big(\frac{L(F(x_1) - F(x^*))}{T_1} + \frac{\mathcal{V}}{|S|}\Big) \leq O\Big(\frac{L\mathcal{V}}{\sigma T_1 T} + \frac{\mathcal{V}T_1}{T}\Big) + \big(1 - \frac{\sigma}{L}\big)^{\Omega(T)}\sigma L\|x_0 - x^*\|^2 \ .$$

After choosing $T_1 \in [1, T]$ to balance the two terms, and noticing $L \geq \sigma$ and $T \geq L/\sigma$, we have

$$\mathbb{E}\Big[\frac{1}{T_1}\sum_{t\in[T_1]}\|\mathcal{G}_{F,\eta}(x_t)\|^2\Big] \leq O\Big(\frac{\sqrt{L}\mathcal{V}}{\sqrt{\sigma}T}\Big) + \big(1 - \frac{\sigma}{L}\big)^{\Omega(T)}L\sigma\|x_0 - x^*\|^2 \ .$$

If we randomly output $x_1, \ldots, x_{T_1}$, we have the desired result. □

**Algorithm 9** `SGD2`$(F, x_0, \sigma, L, T, \eta, T_1)$

**Input:** function $F(x) = \psi(x) + \frac{1}{n}\sum_{i=1}^n f_i(x)$; initial vector $x_0$; parameters $L \geq \sigma > 0$; $T \geq 1$; $T_1 \in [T]$. $\qquad\diamond\ f(x) = \frac{1}{n}\sum_{i=1}^n f_i(x)$ is convex and $L$-smooth
1: $G(x) \stackrel{\text{def}}{=} F(x) + \frac{\sigma}{2}\|x - x_0\|^2$;
2: **return** $\overline{x} \leftarrow \mathtt{SGD1^{sc}}(G, x_0, \sigma, L+\sigma, T, \eta, T_1)$.

## C  Approach 2: SGD After Regularization

In this section, we generalize Nesterov's second trick to the stochastic setting. Namely, we replace $F(x)$ with its regularized version $G(x) = F(x) + \frac{\sigma}{2}\|x - x_0\|^2$ for some small $\sigma > 0$, and then apply our new $\mathtt{SGD1^{sc}}$ method (and Theorem 1(b)) to find a point with vanishing gradient for $G(x)$. This leads to a stationary point for $F(x)$ as long as $\sigma$ is small, owing to Lemma 5.1.

We summarize the result as follows:

**Theorem 2** (SGD2). *Suppose $x^* \in \arg\min_x \{F(x)\}$ and $\sigma \in (0, L/2]$. To solve Problem (3.1) given a starting vector $x_0 \in \mathbb{R}^d$ and $\eta = \frac{C}{L} \leq \frac{1}{8L}$ for some constant $C \leq 1/8$,*

- *If $T_1$ is appropriately chosen, $\mathtt{SGD2}(F, x_0, \sigma, L, T, \eta, T_1)$ finds a point $\overline{x}$ satisfying*
$$\mathbb{E}[\|\mathcal{G}_{F,\eta}(\overline{x})\|^2] \leq O\Big(\sigma^2\|x_0 - x^*\|^2 + \frac{\sqrt{L}\mathcal{V}}{\sqrt{\sigma T}}\Big) + \Big(1 - \frac{\sigma}{L}\Big)^{\Omega(T)} L\sigma\|x_0 - x^*\|^2 \enspace.$$

- *If $\sigma$ is also appropriately chosen, then we find $\overline{x}$ with $\mathbb{E}[\|\mathcal{G}_{F,\eta}(\overline{x})\|^2] \leq \varepsilon^2$ in gradient complexity*
$$T \leq O\Big(\frac{\sqrt{L}\|x_0 - x^*\|\mathcal{V}}{\varepsilon^{2.5}} + \frac{L\|x_0 - x^*\|}{\varepsilon}\log\frac{L\|x_0 - x^*\|}{\varepsilon}\Big) \enspace.$$

(For the same reason as Remark 6.1, the above expected guarantee can be made in high confidence.)

*Proof of Theorem 2.* Let $x_G^*$ be the (unique) minimizer of $G(\cdot)$, which may be different from $x^*$. Applying (1) Theorem 1(b) on $G(x)$, (2) Lemma 5.1 with $S = 1$ and $\widehat{x}_1 = x_0$, and (3) inequality $(a+b)^2 \leq 2a^2 + 2b^2$, we have
$$\mathbb{E}[\|\mathcal{G}_{F,\eta}(\overline{x})\|^2] \leq O\Big(\sigma^2\|x_0 - x_G^*\|^2 + \frac{\sqrt{L}\mathcal{V}}{\sqrt{\sigma T}}\Big) + \Big(1 - \frac{\sigma}{L}\Big)^{\Omega(T)} L\sigma\|x_0 - x_G^*\|^2 \enspace.$$

Now, by definition $\frac{\sigma}{2}\|x^* - x_0\|^2 - \frac{\sigma}{2}\|x_G^* - x_0\|^2 = (G(x^*) - F(x^*)) + (F(x_G^*) - G(x_G^*)) \geq 0$ so we have $\|x_G^* - x_0\| \leq \|x^* - x_0\|$. This proves the first item. The second item is by appropriately tuning $\sigma \in (0, L/2]$. □

## D  Proofs for Section 4

**Theorem 4.1.** *Let $x^* \in \arg\min_x\{F(x)\}$. To solve Problem (3.1) given a starting vector $x_0 \in \mathbb{R}^d$,*

(a) $\mathtt{SGD}(F, x_0, \alpha, T)$ *outputs $\overline{x}$ satisfying $\mathbb{E}[F(\overline{x})] - F(x^*) \leq \frac{\alpha\mathcal{V}}{2(1-\alpha L)} + \frac{\|x_0 - x^*\|^2}{2\alpha T}$ as long as $\alpha < 1/L$. In particular, if $\alpha$ is tuned optimally, it satisfies*
$$\mathbb{E}[F(\overline{x})] - F(x^*) \leq O\Big(\frac{L\|x_0 - x^*\|^2}{T} + \frac{\sqrt{\mathcal{V}}\|x_0 - x^*\|}{\sqrt{T}}\Big) \enspace.$$

(b) *If $f(x)$ is $\sigma$-strongly convex and $T \geq \frac{L}{\sigma}$, then $\mathtt{SGD^{sc}}(F, x_0, \sigma, L, T)$ outputs $\overline{x}$ satisfying*
$$\mathbb{E}[F(\overline{x})] - F(x^*) \leq O\Big(\frac{\mathcal{V}}{\sigma T}\Big) + \Big(1 - \frac{\sigma}{L}\Big)^{\Omega(T)}\sigma\|x_0 - x^*\|^2 \enspace.$$

## D.1 Proof of Theorem 4.1(a)

The following inequality is classically known as the "regret inequality" for proximal mirror descent, and its proof is classical:

**Fact D.1.** *If $x_{t+1} = \arg\min_{y \in \mathbb{R}^d}\{\psi(y) + \frac{1}{2\alpha}\|y - x_t\|^2 + \langle w, y\rangle\}$, then for every $u \in \mathbb{R}^d$:*

$$\langle w, x_{t+1} - u\rangle + \psi(x_{t+1}) - \psi(u) \leq \frac{\|x_t - u\|^2}{2\alpha} - \frac{\|x_{t+1} - u\|^2}{2\alpha} - \frac{\|x_{t+1} - x_t\|^2}{2\alpha} .$$

*Proof.* Recall that the minimality of $x_{t+1} = \arg\min_{y \in \mathbb{R}^d}\{\frac{1}{2\alpha}\|y - x_t\|^2 + \psi(y) + \langle w, y\rangle\}$ implies the existence of some subgradient $g \in \partial \psi(x_{t+1})$ which satisfies $\frac{1}{\alpha}(x_{t+1} - x_t) + w + g = 0$. Combining this with $\psi(u) - \psi(x_{t+1}) \geq \langle g, u - x_{t+1}\rangle$, which is due to the convexity of $\psi(\cdot)$, we immediately have $\psi(u) - \psi(x_{t+1}) + \langle \frac{1}{\alpha}(x_{t+1} - x_t) + w, u - x_{t+1}\rangle \geq \langle \frac{1}{\alpha}(x_{t+1} - x_t) + w + g, u - x_{t+1}\rangle = 0$. Rearranging this inequality we have

$$\langle w, x_{t+1} - u\rangle + \psi(x_{t+1}) - \psi(u) \leq \langle -\frac{1}{\alpha}(x_{t+1} - x_t), x_{t+1} - u\rangle$$

$$= \frac{\|x_t - u\|^2}{2\alpha} - \frac{\|x_{t+1} - u\|^2}{2\alpha} - \frac{\|x_{t+1} - x_t\|^2}{2\alpha} . \qquad \square$$

*Proof of Theorem 4.1(a).* We have the following derivation which is completely classical

$$\mathbb{E}_i[F(x_{t+1}) - F(x^*)] = \mathbb{E}_i[f(x_{t+1}) - f(x^*) + \psi(x_{t+1}) - \psi(x^*)]$$

$$\overset{①}{\leq} \mathbb{E}_i\Big[f(x_t) + \langle \nabla f(x_t), x_{t+1} - x_t\rangle + \frac{L}{2}\|x_t - x_{t+1}\|^2 - f(x^*) + \psi(x_{t+1}) - \psi(x^*)\Big]$$

$$\overset{②}{\leq} \mathbb{E}_i\Big[\langle \nabla f(x_t), x_{t+1} - x_t\rangle + \frac{L}{2}\|x_t - x_{t+1}\|^2 + \langle \nabla f(x_t), x_t - x^*\rangle + \psi(x_{t+1}) - \psi(x^*)\Big]$$

$$= \mathbb{E}_i\Big[\langle \nabla f(x_t), x_{t+1} - x_t\rangle + \frac{L}{2}\|x_t - x_{t+1}\|^2 + \langle \nabla f_i(x_t), x_t - x^*\rangle + \psi(x_{t+1}) - \psi(x^*)\Big]$$

$$= \mathbb{E}_i\Big[\langle \nabla f_i(x_t) - \nabla f(x_t), x_t - x_{t+1}\rangle + \frac{L}{2}\|x_t - x_{t+1}\|^2 + \langle \nabla f_i(x_t), x_{t+1} - x^*\rangle + \psi(x_{t+1}) - \psi(x^*)\Big]$$

$$\overset{③}{\leq} \mathbb{E}_i\Big[\langle \nabla f_i(x_t) - \nabla f(x_t), x_t - x_{t+1}\rangle + \frac{\|x_t - x^*\|^2}{2\alpha} - \frac{\|x_{t+1} - x^*\|^2}{2\alpha} - \Big(\frac{1}{2\alpha} - \frac{L}{2}\Big)\|x_{t+1} - x_t\|^2\Big]$$

$$\overset{④}{\leq} \mathbb{E}_i\Big[\frac{\alpha}{2(1 - \alpha L)}\|\nabla f_i(x_t) - \nabla f(x_t)\|^2 + \frac{\|x_t - x^*\|^2}{2\alpha} - \frac{\|x_{t+1} - x^*\|^2}{2\alpha}\Big] .$$

Above, inequality inequality ① uses the fact that $f(\cdot)$ is $L$-smooth; inequality ② uses the convexity of $f(\cdot)$; inequality ③ uses Fact D.1 and inequality ④ uses Young's inequality $\langle a, b\rangle \leq \frac{1}{2}\|a\|^2 + \frac{1}{2}\|b\|^2$. Next, we telescope the above inequality for $t = 0, 1, \ldots, T-1$ and use $\mathbb{E}_i[\|\nabla f_i(x) - \nabla f(x)\|^2] \leq \mathcal{V}$:

$$\mathbb{E}\Big[\sum_{t=1}^T (F(x_t) - F(x^*))\Big] \leq \frac{\alpha \mathcal{V} \cdot T}{2(1 - \alpha L)} + \frac{\|x_0 - x^*\|^2}{2\alpha} .$$

Therefore, if we choose $\alpha \leq \frac{1}{2L}$ to balance the two terms, we have $\overline{x} = \frac{1}{T}(x_1 + \cdots + x_T)$ satisfies

$$\mathbb{E}[F(\overline{x})] - F(x^*) \leq O\Big(\frac{L\|x_0 - x^*\|^2}{T} + \frac{\sqrt{\mathcal{V}}\|x_0 - x^*\|}{\sqrt{T}}\Big) . \qquad \square$$

## D.2 Proof of Theorem 4.1(b)

*Proof of Theorem 4.1(b).* Since $f(x)$ is $\sigma$-strongly convex, the proof of Theorem 4.1(a) tells us by applying SGD once for $T$ iterations, we can obtain a point, denoted by $x_1$, satisfying

$$\mathbb{E}[F(x_1)] - F(x^*) \leq \frac{\alpha \mathcal{V}}{2(1-\alpha L)} + \frac{\|x_0 - x^*\|^2}{2\alpha T} \leq \frac{\alpha \mathcal{V}}{2(1-\alpha L)} + \frac{F(x_0) - F(x^*)}{\sigma \alpha T} \ . \tag{D.1}$$

Now, following the idea of [19], we repeatedly apply Theorem 4.1(a) to get the tightest result.

In particular, we first apply (D.1) for $N = \lfloor \frac{T}{8L/\sigma} \rfloor$ rounds, each with $\alpha_k = 1/2L$ and $T_k = 4L/\sigma$. By induction, (D.1) ensures that we can obtian a point $x_N$ satisfying

$$\mathbb{E}[F(x_N)] - F(x^*) \leq \frac{\mathcal{V}}{L} + \left(1 - \frac{\sigma}{L}\right)^{\Omega(T)} \sigma \|x_0 - x^*\|^2 \ .$$

Next, we apply (D.1) for $K = \lfloor \log_2(\sigma T/16L) \rfloor$ additional rounds, $k = 1, 2, \ldots, K$, each time with $T_k = 2^{k+2} \frac{L}{\sigma}$ and $\alpha_k = \frac{1}{2^k L}$. Again, by induction, (D.1) implies

$$\mathbb{E}[F(x_{N+K})] - F(x^*) \leq O\left(\frac{\mathcal{V}}{2^K L}\right) + \left(1 - \frac{\sigma}{L}\right)^{\Omega(T)} \sigma \|x_0 - x^*\|^2 \ .$$

Finally, notice that the total gradient complexity is at most $N \cdot \frac{4L}{\sigma} + 2^{K+3} \frac{L}{\sigma} \leq \frac{T}{2} + \frac{T}{2} = T$ and $T = \Theta(2^K L/\sigma)$. This finishes the proof. □

## E  Missing Proof for SGD5

In Section E.1, we review Oja's algorithm which is an online method for finding eigenvectors. In Section E.2, we state some simple auxiliary claims.

### E.1 Oja's Algorithm

Let $\mathcal{D}$ be a distribution over $d \times d$ symmetric matrices whose eigenvalues are between 0 and 1, and denote by $\mathbf{B} \stackrel{\text{def}}{=} \mathbb{E}_{\mathbf{A} \sim \mathcal{D}}[\mathbf{A}]$ its mean. Let $\mathbf{A}_1, \ldots, \mathbf{A}_T$ be $T$ copies of i.i.d. samples generated from $\mathcal{D}$. Oja's algorithm begins with a random unit-norm Gaussian vector $w_1 \in \mathbb{R}^d$. At each iteration $k \in 2, \ldots, T$, Oja's algorithm computes $w_k = \frac{(\mathbf{I}+\eta \mathbf{A}_{k-1}) w_{k-1}}{C}$ where $C > 0$ is the normalization constant such that $\|w_k\| = 1$. Allen-Zhu and Li [5] showed (see its last section) that [13]

**Theorem E.1.** *For every $p \in (0,1)$, choosing $\eta = \Theta(\sqrt{p/T})$, we have with prob. $\geq 1-p$:*

$$\sum_{k=1}^T w_k^\top \mathbf{B} w_k \geq T \cdot \lambda_{\max}(\mathbf{B}) - O\left(\frac{\sqrt{T}}{\sqrt{p}} \cdot \log(d/p)\right) \ .$$

*Remark* E.2. The above result is asymptotically optimal even in terms of sampling complexity [5].

**Approximating MinEV of Hessian.** Suppose $f(x) = \frac{1}{n} \sum_{i=1}^n f_i(x)$ where each $f_i(x)$ is twice-differentiable and $L$-smooth. We can denote by $\mathcal{D}$ the distribution where each $\frac{L \cdot \mathbf{I} - \nabla^2 f_i(x)}{2L} \in \mathbb{R}^{d \times d}$ is generated with probability $\frac{1}{n}$, and then use Oja's algorithm to compute the minimum eigenvalue of $\nabla^2 f(x)$. Note each time when computing $(\mathbf{I} + \eta \mathbf{A}_{k-1}) w_{k-1}$, it suffices to compute Hessian-vector product (i.e., $\nabla^2 f_i(x) \cdot w_{k-1}$) once. The following corollary is simple to prove (see for instance [3]):

**Lemma E.3.** *There exists absolute constant $C > 1$ such that for any $x \in \mathbb{R}^d$, $T \geq 1$, $p \in (0,1)$:*

---
[13] The original one-paged proof from [5] only showed Theorem E.1 where the left hand side is $\sum_{k=1}^T w_k^\top \mathbf{A}_k w_k$. However, by Azuma's inequality, we have $\sum_{k=1}^T w_k^\top \mathbf{B} w_k \geq \sum_{k=1}^T w_k^\top \mathbf{A}_k w_k - O(\sqrt{T \log(1/p)})$ with probability $\geq 1-p$.

- if we run Oja's algorithm once for $T$ iterations, with $\eta = \Theta(\sqrt{T})$, we can find unit vector $y$ such that, with at with probability at least $4/5$,
$$y^\top \nabla^2 f(x) y \leq \lambda_{\min}(\nabla^2 f(x)) + C \cdot \frac{L \log(d)}{\sqrt{T}} \ .$$

- if we run Oja's algorithm $O(\log(1/p))$ times each for $T$ iterations, then w.p. $\geq 1 - p$, we can
$$\text{either conclude} \quad \lambda_{\min}(\nabla^2 f(x)) \geq -C \cdot \frac{L \log(d/p)}{\sqrt{T}} \ ,$$
$$\text{or find } y \in \mathbb{R}^d: \quad y^\top \nabla^2 f(x) y \leq -\frac{C}{2} \cdot \frac{L \log(d/p)}{\sqrt{T}} \ .$$
The total number of Hessian-vector products is at most $O(T \log(1/p))$.

When $f_i(x)$ is explicitly given, the computational complexity for computing a Hessian-vector product $\nabla^2 f_i(x) \cdot v$ is roughly twice that for computing $\nabla f_i(x)$. For such reason, we usually denote gradient complexity $T$ as the total number of computations of stochastic gradients *plus* those of Hessian-vector products. If one insists in forbidding Hessian-vector product computations, the result [6] also designed a variant of Oja's algorithm which achieves the same guarantee as Lemma E.3 but using only stochastic gradient computations (without Hessian-vector products).

### E.2 Auxiliary Claims

The following claim is a simple consequence of smoothness definition, see [10, Lemma 4.1].

**Claim E.4** ([10]). *If $f(x)$ is $L$-smooth and $L_2$-second-order smooth, and $y \in \mathbb{R}^d$ is a point such that $\nabla^2 f(y) \succeq -\delta \mathbf{I}$ for some $\delta > 0$, then the function $F(x) = f(x) + L\big(\max\{0, \|x - y\| - \frac{\delta}{L_2}\}\big)^2$ is $5L$-smooth and $3\delta$-nonconvex.*

The following claim is also a consequence of smoothness definition.

**Claim E.5** ([3]). *If $v$ is a unit vector with $v^\top \nabla^2 f(y_k) v \leq -\frac{\delta}{2}$ and $y_{k+1} = y_k \pm \frac{\delta}{L_2} v$ for a random sign, then $f(y_k) - \mathbb{E}[f(y_{k+1})] \geq \frac{\delta^3}{12 L_2^2}$.*

*Proof.* Suppose $y_{k+1} = y_k \pm \eta v$ where $\|v\| = 1$ and $\eta = \frac{\delta}{L_2}$, then by the second-order smoothness,
$$f(y_k) - \mathbb{E}[f(y_{k+1})] \geq \mathbb{E}\Big[\langle \nabla f(y_k), y_k - y_{k+1}\rangle - \frac{1}{2}(y_k - y_{k+1})^\top \nabla^2 f(y_k)(y_k - y_{k+1}) - \frac{L_2}{6}\|y_k - y_{k+1}\|^3\Big]$$
$$= -\frac{\eta^2}{2} v^\top \nabla^2 f(y_k) v - \frac{L_2 \eta^3}{6}\|v\|^3 \geq \frac{\eta^2 \delta}{4} - \frac{L_2 \eta^3}{6} = \frac{\delta^3}{12 L_2^2} \ . \qquad \square$$

**Claim E.6.** *In each iteration $k$ of `SGD5`, we have*
$$f(y_k) - \mathbb{E}[f(y_{k+1})] \geq \Omega(1) \cdot \mathbb{E}\Big[\widetilde{\sigma}\|y_k - y_{k+1}\|^2 + \widetilde{\sigma}\|y_k - y_k^*\|^2\Big] - O\Big(\frac{\mathcal{V}}{\widetilde{\sigma} B}\Big) \ ,$$
*where $y_k^* \stackrel{\text{def}}{=} \arg\min_x \{G^k(x)\} = \arg\min_x\{F^k(x) + \widetilde{\sigma}\|x - y_k\|^2\}$.*

*Proof.* By directly applying (7.2) (note that we can do so because $B = \Theta(\frac{\mathcal{V}}{\varepsilon^2}) \geq \Omega(\frac{\widetilde{L}}{\widetilde{\sigma}})$), we have
$$\mathbb{E}\Big[\widetilde{\sigma}\|y_k - y_{k+1}\|^2 + \widetilde{\sigma}\|y_k - y_k^*\|^2\Big] \leq \mathbb{E}\Big[F^k(y_k) - F^k(y_{k+1})\Big] + O\Big(\frac{\mathcal{V}}{\widetilde{\sigma} B}\Big) \ .$$
Noting that $F^k(y_k) = f(y_k)$ but $F^k(y_{k+1}) \geq f(y_{k+1})$, we finish the proof. $\qquad \square$

### E.3 Proof of Theorem 5

Throughout the proof of Theorem 5, we shall use the big-$\Theta$ notion to hide absolute constants, in order to simplify notations.

*Proof of Theorem 5.* Recall $N_1 = \Theta\big(\frac{\widetilde{\sigma}\Delta_f}{\varepsilon^2}\big)$ is the number of first-order steps. We denote by $N_2$ the actual number of second-order steps, which is a *random variable*.

We first note that each call of Oja's algorithm succeeds with probability at least $1 - \frac{1}{20(k+1)^2}$, and therefore by $\sum_{k=1}^{\infty} k^{-2} < 1.65$, with probability at least $1 - \frac{1}{12}$ (over the randomness of Oja's algorithm only), all occurrences of Oja's algorithm succeed. In the remainder of the proof, we shall always assume that this event happens. In other words, in Line 6 of SGD5, it either finds $v^\top \nabla^2 f(y_k) v \leq -\frac{\delta}{2}$ or if not, conclude that $\nabla^2 f(y_k) \succeq -\delta \mathbf{I}$. (Recall Lemma E.3.)

Let us define random variables $\Delta_1, \Delta_2$ the total amount of objective decrease during first-order and second-order steps respectively.[14] By Claim E.6 and the fact that there are exactly $N_1$ first-order steps, we have $\mathbb{E}[\Delta_1] \geq -O(\frac{\mathcal{V}}{\widetilde{\sigma}B}) \cdot N_1 = -O(\frac{\mathcal{V}}{B\varepsilon^2}) \cdot \Delta_f \geq -\Delta_f$, where the last inequality is due to our choice of $B$.

**Accuracy.** Since $\Delta_1 + \Delta_2 \leq \Delta_f$ and $\mathbb{E}[\Delta_2] \geq 0$ by Claim E.5, we conclude that if we select $k = 0, 1, \ldots,$ at random among the $N_1$ first-order steps, then

$$\mathbb{E}[f(y_k) - f(y_{k+1})] \leq \frac{\mathbb{E}[\Delta_1]}{N_1} \leq \frac{\Delta_f - \mathbb{E}[\Delta_2]}{N_1} \leq \frac{\Delta_f}{N_1} \ .$$

Denote by $y = y_k$, $y^+ = y_{k+1}$, and $y^* = \arg\min_x\{G^k(x)\}$ for this random choice of $k$. Combining $\mathbb{E}[f(y_k) - f(y_{k+1})] \leq \frac{\Delta_f}{N_1}$ and Claim E.6, we have

$$\mathbb{E}\Big[\widetilde{\sigma}\|y - y^+\|^2 + \widetilde{\sigma}\|y - y^*\|^2\Big] \leq O\Big(\frac{\Delta_f}{N_1} + \frac{\mathcal{V}}{\widetilde{\sigma}B}\Big) = O\Big(\frac{\varepsilon^2}{\widetilde{\sigma}}\Big) \ .$$

By Markov's bound, with probability at least, $1 - \frac{1}{12}$, we have

$$\widetilde{\sigma}\|y - y^+\|^2 + \widetilde{\sigma}\|y - y^*\|^2 \leq O\Big(\frac{\varepsilon^2}{\widetilde{\sigma}}\Big) \ . \tag{E.1}$$

Now, recall that

$$F(x) \stackrel{\text{def}}{=} f(x) + L(\max\{0, \|x - y\| - \tfrac{\delta}{L_2}\})^2 \quad \text{and} \quad G(x) = F(x) + \widetilde{\sigma}\|x - y\|^2$$

we can apply SGD3$^{\text{sc}}$ with gradient complexity $T_{\text{sgd}}$ to minimize $G(x)$. Let the output be $x^{\text{out}}$. Using Theorem 3(a) (with $T = T_{\text{sgd}}$), we have with probability at least $1 - \frac{1}{12}$

$$\|\nabla G(x^{\text{out}})\|^2 \leq O\Big(\frac{\mathcal{V} \cdot \log^3 \frac{\widetilde{L}}{\widetilde{\sigma}}}{T_{\text{sgd}}}\Big) + \big(1 - \frac{\widetilde{\sigma}}{\widetilde{L}}\big)^{\Omega(T_{\text{sgd}}/\log(\widetilde{L}/\widetilde{\sigma}))} \widetilde{\sigma}^2 \|y - y^*\|^2 \tag{E.2}$$

Using Lemma 7.2, we have

$$\|\nabla F(x^{\text{out}})\|^2 + \widetilde{\sigma}^2 \|x^{\text{out}} - y\|^2 \leq O\big(\widetilde{\sigma}^2 \|y - y^*\|^2 + \|\nabla G(x^{\text{out}})\|^2\big) \ . \tag{E.3}$$

Combining (E.1), (E.2), and (E.3), and our choice $T_{\text{sgd}} = \Theta\big(\frac{\mathcal{V}}{\varepsilon^2} \log^3 \frac{\widetilde{L}}{\widetilde{\sigma}}\big) \geq \Omega\big(\frac{\widetilde{L}}{\widetilde{\sigma}} \log \frac{\widetilde{L}}{\widetilde{\sigma}}\big)$ we have

$$\|\nabla F(x^{\text{out}})\|^2 + \widetilde{\sigma}^2 \|x^{\text{out}} - y\|^2 \leq O\Big(\varepsilon^2 + \frac{\mathcal{V} \cdot \log^3 \frac{\widetilde{L}}{\widetilde{\sigma}}}{T_{\text{sgd}}}\Big) \leq \varepsilon^2 \ . \tag{E.4}$$

---
[14] More precisely, $\Delta_1 \stackrel{\text{def}}{=} \sum_{k=0}^{\infty} \mathbb{I}[\text{iter } k \text{ exists and is a first-order step}] \cdot (f(y_k) - f(y_{k+1}))$ , and similarly for $\Delta_2$.

Since we choose parameter $\widetilde{\sigma}$ so that $\frac{\varepsilon}{\widetilde{\sigma}} \leq O(\frac{\delta}{L_2})$, (E.4) implies

$$\|x^{\mathsf{out}} - y\| \leq \frac{\varepsilon}{\widetilde{\sigma}} \leq \frac{\delta}{L_2} \ .$$

In other words, $x^{\mathsf{out}}$ is not too far away from $y$ and therefore by definition $F(x) \stackrel{\text{def}}{=} f(x) + L(\max\{0, \|x - y\| - \frac{\delta}{L_2}\})^2$,

$$\nabla^2 F(x^{\mathsf{out}}) = \nabla^2 f(x^{\mathsf{out}}) \quad \text{and} \quad \nabla F(x^{\mathsf{out}}) = \nabla f(x^{\mathsf{out}}) \ .$$

This means $\nabla^2 f(x^{\mathsf{out}}) = \nabla^2 F(x^{\mathsf{out}}) \succeq -3\delta \mathbf{I}$ since $F(\cdot)$ is $3\delta$-nonconvex (see Claim E.4) and $\|\nabla f(x^{\mathsf{out}})\| = \|\nabla F(x^{\mathsf{out}})\| \leq \varepsilon$ by (E.4). This finishes the proof of the accuracy of SGD5.

**Running Time.** Recall that random variable $N_2$ is the number of second-order steps. By Claim E.5, we have

$$\mathbb{E}[N_2] \cdot \frac{\delta^3}{12 L_2^2} \leq \mathbb{E}[\Delta_2] \leq \Delta_f - \mathbb{E}[\Delta_1] \leq 2\Delta_f \implies \mathbb{E}[N_2] \leq O\Big(\frac{L_2^2 \Delta_f}{\delta^3}\Big) \ .$$

Therefore, with probability at least $1 - \frac{11}{12}$, we have $N_2 \leq O\big(\frac{L_2^2 \Delta_f}{\delta^3}\big)$. The remainder of the derivation always assumes this event happens.

The total gradient complexity $T$ consists of three parts:

- The gradient complexity for Oja's algorithms is at most $O\big((N_1 + N_2) \frac{L^2}{\delta^2}\big)$.
- The gradient complexity for SGD$^{\mathsf{sc}}$ for $N_1 = \Theta\big(\frac{\widetilde{\sigma} \Delta_f}{\varepsilon^2}\big)$ times is at most $N_1 \cdot B$.
- The gradient complexity for SGD3$^{\mathsf{sc}}$ in the end is $T_{\mathsf{sgd}} = \widetilde{O}\big(\frac{\mathcal{V}}{\varepsilon^2}\big)$.

**Case 1.** Suppose $L_2 \geq \frac{L\delta}{\varepsilon}$. This corresponds to the case when $L_2$ is too large. Recall we have chosen $\widetilde{L} = \widetilde{\sigma} = \Theta(\frac{\varepsilon L_2}{\delta})$ and $N_1 = \Theta\big(\frac{\widetilde{\sigma} \Delta_f}{\varepsilon^2}\big)$. The total gradient complexity is

$$T = \widetilde{O}\Big(T_{\mathsf{sgd}} + (N_1 + N_2) \cdot \frac{L^2}{\delta^2} + N_1 \cdot \frac{\mathcal{V}}{\varepsilon^2}\Big) \leq \widetilde{O}\Big(\frac{\mathcal{V}}{\varepsilon^2} + \Big(\frac{\widetilde{L}\Delta_f}{\varepsilon^2} + \frac{L_2^2 \Delta_f}{\delta^3}\Big) \cdot \frac{L^2}{\delta^2} + \frac{\widetilde{L}\Delta_f}{\varepsilon^2} \cdot \frac{\mathcal{V}}{\varepsilon^2}\Big)$$
$$\leq \widetilde{O}\Big(\frac{\mathcal{V}}{\varepsilon^2} + \frac{L_2^2 \Delta_f}{\delta^3} \cdot \frac{L^2}{\delta^2} + \frac{L_2 \Delta_f}{\varepsilon \delta} \cdot \frac{\mathcal{V}}{\varepsilon^2}\Big) \ .$$

**Case 2.** Suppose $L_2 \leq \frac{L\delta}{\mathcal{V}^{1/3} \varepsilon^{1/3}}$. This is the *interesting case* and recall we have chosen $\widetilde{L} = L$, and $\widetilde{\sigma}$ is large enough so that $\widetilde{\sigma} = \Omega\big(\max\big\{\frac{\varepsilon L_2}{\delta}, \frac{\varepsilon^2 L}{\mathcal{V}}\big\}\big)$. (It is easy to verify that this value $\widetilde{\sigma}$ is no greater than $L$.) The total gradient complexity

$$T = \widetilde{O}\Big(T_{\mathsf{sgd}} + (N_1 + N_2) \cdot \frac{L^2}{\delta^2} + N_1 \cdot \frac{\mathcal{V}}{\varepsilon^2}\Big) \leq \widetilde{O}\Big(\frac{\mathcal{V}}{\varepsilon^2} + \Big(\frac{\widetilde{\sigma} \Delta_f}{\varepsilon^2} + \frac{L_2^2 \Delta_f}{\delta^3}\Big) \cdot \frac{L^2}{\delta^2} + \frac{\widetilde{\sigma} \Delta_f}{\varepsilon^2} \cdot \frac{\mathcal{V}}{\varepsilon^2}\Big)$$
$$= \widetilde{O}\Big(\frac{\mathcal{V}}{\varepsilon^2} + \frac{L_2^2 L^2 \Delta_f}{\delta^5} + \frac{\widetilde{\sigma} \Delta_f}{\varepsilon^2} \cdot \Big(\frac{L^2}{\delta^2} + \frac{\mathcal{V}}{\varepsilon^2}\Big)\Big) = \widetilde{O}\Big(\frac{\mathcal{V}}{\varepsilon^2} + \frac{L_2^2 L^2 \Delta_f}{\delta^5} + \frac{L_2 \Delta_f}{\varepsilon \delta} \cdot \frac{\mathcal{V}}{\varepsilon^2} + \frac{L \Delta_f}{\mathcal{V}} \cdot \frac{L^2}{\delta^2}\Big) \ .$$

$\square$